\documentclass[pdflatex,sn-mathphys-num]{sn-jnl}

\usepackage{amsmath,amssymb,amsfonts}
\usepackage{amsthm}
\usepackage{mathrsfs}

\usepackage{graphicx}
\usepackage[table]{xcolor}
\usepackage{booktabs}
\usepackage{multirow}
\usepackage{siunitx}
\usepackage{adjustbox}
\usepackage{threeparttable}
\usepackage{float}    
\usepackage{placeins}
\usepackage{stfloats}  
\usepackage{cuted}      
\usepackage{pifont}
\usepackage{subcaption} 
\usepackage[labelfont=bf]{caption}

\usepackage{xcolor}
\usepackage{textcomp}
\usepackage{manyfoot}
\usepackage{algorithm}
\usepackage{algorithmicx}
\usepackage{algpseudocode}
\usepackage{listings}
\usepackage[title]{appendix}
\usepackage{verbatim}
\usepackage{tipa}
\usepackage{url}
\usepackage{natbib}

\usepackage{hyperref}

\hypersetup{colorlinks=true,allcolors=black}
\theoremstyle{thmstyleone}

\theoremstyle{thmstyletwo}

\theoremstyle{thmstylethree}

\newcounter{suppfigcounter}

\newcommand{\suppfigsection}[2]{%
    \refstepcounter{suppfigcounter}
    \subsection*{Supplementary Figure \thesuppfigcounter: #1}%
    \label{#2}%
}
\expandafter\ifx\csname pdfobjcompresslevel\endcsname\relax\else
\fi
\begin{document}

\title[Exploiting heterogeneous delays for efficient computation in low-bit neural networks]{Exploiting heterogeneous delays for efficient computation in low-bit neural networks}

\author*[1]{\fnm{Pengfei} \sur{Sun}} \email{p.sun@imperial.ac.uk} 
\author[2]{\fnm{Jascha} \sur{Achterberg}}
\author[3]{\fnm{Zhe} \sur{Su}}
\author[1]{\fnm{Dan F.M.} \sur{Goodman}} 
\author*[1,4,5]{\fnm{Danyal} \sur{Akarca}}\email{d.akarca@imperial.ac.uk}
\affil[1]{\orgdiv{Department of Electrical and Electronic Engineering}, \orgname{Imperial College London}}
\affil[2]{\orgdiv{Centre for Neural Circuits and Behaviour}, \orgname{University of Oxford}}
\affil[3]{\orgdiv{
Institute of Neuroinformatics}, \orgname{University of Zurich and ETH Zurich}}
\affil[4]{\orgdiv{Imperial-X}, \orgname{Imperial College London}}
\affil[5]{\orgdiv{MRC Cognition and Brain Sciences Unit}, \orgname{University of Cambridge}}

\abstract{
Neural networks rely on learning synaptic weights. However, this overlooks other learnable neural parameters. One such parameter is delay: neurons communicate asynchronously with heterogeneous transmission delays. We propose that, rather than a cost to minimize, this heterogeneity supports computation when task-relevant information unfolds over time. We test this hypothesis by jointly learning weights and delays at different precisions in feedforward spiking neural networks. Delay heterogeneity supports competitive performance on neuromorphic benchmarks while compressing weights over an order of magnitude more aggressively than typically studied weight-only networks. Performance depends on task-matched delay distributions, with temporally complex tasks selectively relying on longer delays. These benefits persist under explicit delay constraints, while neural time constants only partially substitute for long delays. Our results connect the temporal structure of data with efficient model design, highlighting temporal heterogeneity as a computational resource with implications for embodied intelligence and neuromorphic hardware.}

\maketitle

\section{Introduction}
The brain has inherent time delays across all scales of its organization \citep{sreenivasan2019and}. {In engineered computing systems, and in much of contemporary machine-learning practice built around highly synchronized operations,} the dominant assumption has been that these delays exist simply as a necessary byproduct of being a physical system, providing limiting constraints on neural computation and communication \citep{keller2024spacetimeperspectivedynamicalcomputation}. This assumption is reflected in artificial neural networks, the prevailing model of neural computation \citep{https://doi.org/10.1111/cogs.12148,doerig2023neuroconnectionist}, which rely on learning and deploying synaptic weights through highly synchronized operations that are explicitly optimized to minimize delays \citep{volkov2016understanding}. It is hard to imagine when this assumption would not be correct, particularly when considering \textit{individual} operations or signals being transmitted and received between two neurons.

However, when considering \textit{systems} of operations or signals that must be coordinated, several lines of evidence challenge this assumption. 
{A systems-level perspective from neuroscience and nonlinear dynamics suggests that coupling delays can shape collective dynamics and modulate communication between neural populations \citep{bastos2015communication,petkoski2016heterogeneity,petkoski2019transmission}. Related works also appear in delay-based dynamical substrates such as delay-reservoir computing, where fixed delayed feedback can emulate high-dimensional temporal states \citep{appeltant2011single,hart2019delayed,soriano2014delay,ortin2017reservoir}. In most of this work, delays are treated as fixed design or structural parameters rather than learned end-to-end from task data, leaving open how {learnable transmission delays} complement synaptic learning under explicit resource constraints. Complementing these perspectives, work in computational neuroscience indicates that heterogeneity in intrinsic neural time scales can improve robustness and support richer temporal computations on structured tasks \citep{perez2021neural,gast2024heterogeneity}.} 

Biologically, recent evidence shows the brain actively orchestrates temporal signals, from retinal circuits organizing precise timing through circuit and conduction constraints \citep{bucci2025synchronization} to activity-dependent myelination playing a previously underappreciated role in learning by fine-tuning transmission delays \citep{bonetto2021myelin}, with delays also implicated in shaping collective synchronization at the network level \citep{petkoski2016heterogeneity,petkoski2019transmission}. Theoretically, delays enable memory storage without recurrent architectures \citep{izhikevich2006polychronization}, expand computable functions \citep{maass1999complexity} and allow temporal multiplexing for computational resolution through timing rather than hardware scaling \citep{edwards2024computingclocks}. Computationally, temporal heterogeneity as observed in the brain's distribution of slow to fast time constants \cite{hawrylycz2012anatomically,manis2019classification}, improves efficiency for robust learning in dynamic environments \citep{perez2021neural}. This convergent biological, theoretical, and computational evidence prompts a fundamental question: Can the brain's delay heterogeneity be \textit{exploited} to provide complementary benefits beyond learning neural weights alone? If true, this points to a deeper principle whereby systems can harness their inherent structure for flexible, asynchronous, task-adaptive computation.

{We hypothesize that delay heterogeneity confers an advantage on temporally structured tasks by aligning external temporal structure with a network’s internal degrees of freedom. We quantify when temporal heterogeneity can substitute for synaptic precision under a fixed parameter-memory budget. Here, temporal parameters provide an additional representational resource that can be allocated alongside synaptic precision (in bits \cite{shannon1948mathematical}). We test this hypothesis in feedforward spiking neural networks (SNNs) that jointly optimise synaptic weights and axonal delays. Across neuromorphic tasks with increasing temporal structure, we show four consistent findings. First, learning delays yields strong gains with a parameter-efficient parameterisation, and supports a simple fine-tuning regime in which delays improve weight-trained networks \citep{shrestha2018slayer,hammouamri2024learning,goltz2025delgrad, meszaros2025efficient,sun2025algorithm}. Second, delay heterogeneity exposes a task-dependent ``bit-budget'' regime in which temporal precision can compensate for highly imprecise weights (as low as 1.58-bit precision), retaining accuracy close to strong full-precision baselines even when weights are compressed over an order of magnitude more aggressively than weight-only baselines \citep{shuvaev2024encoding}. We further show that, in this regime, the gains arise from temporal heterogeneity itself rather than from delay optimisation alone: fixing heterogeneous delays sampled from simple priors and training only the weights already yields measurable recovery under extreme quantisation, whereas learning the delays shapes a task-matched distribution and delivers the strongest improvements.  Third, systematic ablations show that delay contributions are structured rather than uniform: temporally complex tasks exhibit selective reliance on longer delays, whereas less temporally structured tasks show weaker sensitivity to long delays \citep{betzel2018specificity}. Finally, we characterise how delay heterogeneity remains effective under explicit delay constraints, with neural time constants only partially substituting for long delays, jointly shaping the resource efficiency of the learned solutions.}

We suggest temporal heterogeneity, here reflected through heterogeneous delays, reflects a fundamental computational principle for embodied artificial intelligence that unlocks new modes of computation beyond weights alone.

\section{Results}\label{sec2}
\begin{figure}[!tbp]
\centering
\includegraphics[width=\linewidth]{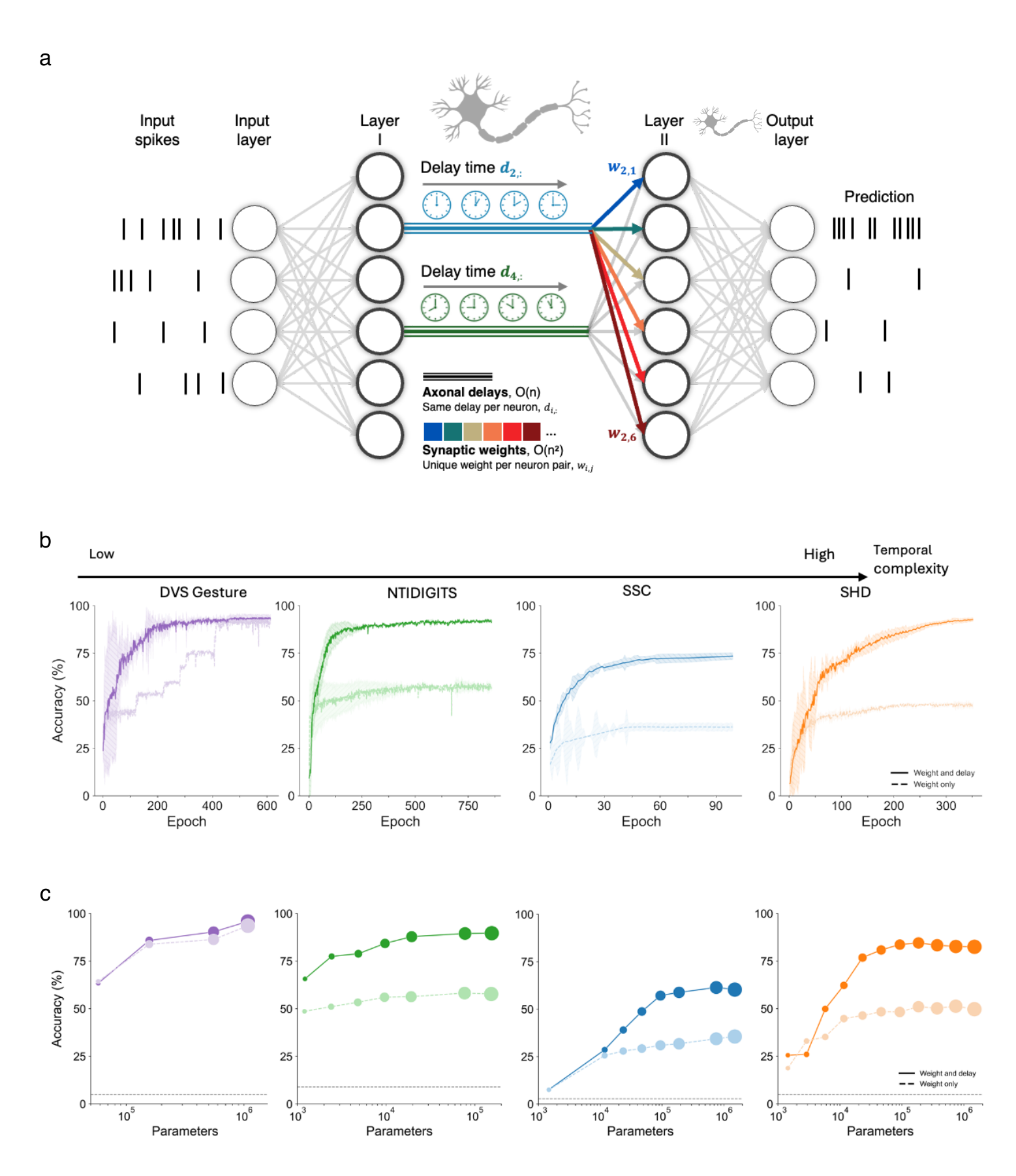}
\caption{\textbf{Training delays provides parameter-efficient performance improvements according to task temporal complexity.}
\textbf{a}. Each neuron maintains one learned delay parameter shared across all outgoing connections ($O(n)$ scaling), while synaptic weights require individual parameters for each connection ($O(n^2)$ scaling). Circles represent neurons, lines represent connections. 
\textbf{b}. Classification accuracy learning curve across four neuromorphic datasets comparing networks with weights only (dashed line) versus weights and axonal delays (solid line). Performance deviation represents standard error across $n=10$ independent runs. Delays consistently improve performance.
\textbf{c}. Per-parameter performance analysis shows the efficiency advantage of delays relative to weights. The y-axis shows accuracy  per  parameter. Delays (solid) achieve significantly greater performance gains per parameter than equivalent weight additions (dashed) across all tasks. Circle size indicates neuron count ($n=16\times2^{k}$, up to $2048$).
}
\label{fig1}
\end{figure}

\subsection{Delay heterogeneity provides disproportionate performance gains over weights on temporal tasks}

We investigated delay heterogeneity in SNNs by training both synaptic weights and axonal delays to classify auditory and visual stimuli with varying temporal structure \citep{perez2021neural}. We implemented axonal delays as our delay learning mechanism because this is the most computationally tractable approach, where each neuron maintains only one learned delay parameter shared across all outgoing connections. This results in $O(n)$ parameter scaling compared to the $O(n^2)$ scaling of connection weights (\textbf{Figure \ref{fig1}a}). 

We tested on four datasets spanning temporal complexity: Spiking Heidelberg Digits (SHD), Spiking Speech Commands (SSC) \citep{cramer2020heidelberg}, Neuromorphic TIDIGITS (NTIDIGITS) \citep{anumula2018feature}, and DVS Gesture \citep{Amir2017}. SHD, SSC and NTIDIGITS all consist of auditory stimuli transformed into spike trains by simplified models of the auditory nerve (NTIDIGITS) and the cochlear nucleus (SHD, SSC), the first and second obligatory relays in the auditory periphery.
DVS Gesture employs event‑driven vision sensors to encode hand gestures as spikes. The first three datasets contain strong temporal structure requiring temporal integration for good performance, while DVS Gesture can be seen to have relatively little temporal structure, as collapsing all events into a single frame preserves a substantial fraction of classification-relevant information. For audio tasks, our model used four layers: an input layer encoding stimuli, two hidden feedforward layers with trainable neuron-specific axonal delays, and a readout layer producing spike-based predictions. For the visual DVS Gesture, we used two spiking convolutional layers plus one fully-connected layer, a standard minimal architecture from prior work \citep{shrestha2018slayer,dampfhoffer2022investigating,sun2025towards}.

We found that for every task studied, learning axonal delays in addition to neural weights significantly increased peak attained task performance, corroborating recent findings pointing to the benefit of delays, achieving results commensurate with state-of-the-art performance \citep{sun2022axonal,sun2024delay,nowotny2025loss,10757311,hammouamri2024learning,queant2025delrec,bouhadjar2026sparse} (\textbf{Figure \ref{fig1}b}). \textbf{Supplementary Figure \ref{sup1}} shows resultant weight and delay distributions over the course of training, broken down by task. Despite broad improvement across tasks, delays had differential task-specific benefits: on SHD, delays improved accuracy from 48.60\% to 90.98\%. Similar substantial gains were observed on SSC (from 38.50\% to 70.94\%) and NTIDIGITS (from 60.02\% to 93.28\%) while on DVS Gesture the improvement was minimal, from 93.56\% to 95.83\% (\textbf{Supplementary Table~{\ref{tab:accuracy}}}), because the reduced temporal complexity means it is possible to do well at this task without taking time into account (in accordance with earlier findings \citep{perez2021neural}).

While axonal delays scale more efficiently than weights ($O(n)$ versus $O(n^2)$), the combined weight-delay networks still contain additional parameters overall. As such, we analyzed performance gains on a per-parameter basis, revealing that delays deliver significantly greater efficiency improvements than equivalent weight additions (\textbf{Figure \ref{fig1}c}). Moreover, in \textbf{Supplementary Figure \ref{sup2}}, we show that these findings do not necessarily require training both weights and delays together for the whole course of training. Instead, only a small amount of weight training is required before delay training can entirely take over: meaning that spiking networks with weights only can be fine-tuned with delays to attain comparable peak performance.

Together, these results demonstrate that axonal delays provide a parameter-efficient mechanism for capturing temporal dependencies, with benefits scaling directly with task complexity with large gains on temporally demanding benchmarks and negligible effects on simpler datasets \citep{sun2022axonal,10181778,zhang2020supervised,yu2022improving,sun2024delay,hammouamri2024learning,d2024denram,zheng2024temporal}. Moreover, delay-augmented feedforward networks support longer effective timescales and achieve competitive performance with substantially lower parameter memory than recurrent spiking architectures\citep{sun2025algorithm,cramer2020heidelberg,yan2025efficient,yin2021accurate,pagkalos2023introducing}.

\begin{figure*}[]
\centering
\includegraphics[width=1\linewidth]{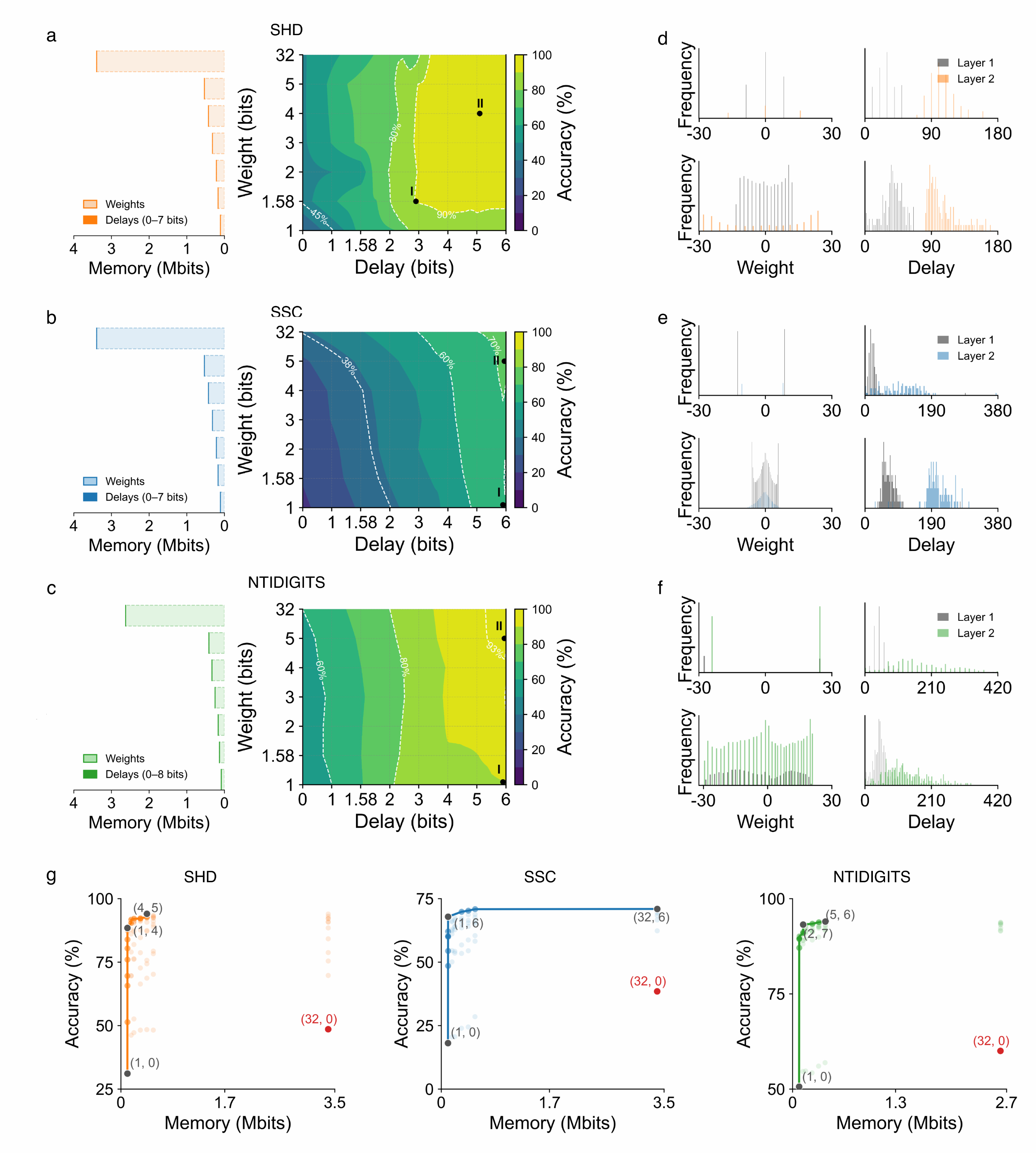}
\caption{\textbf{Weight-delay trade-offs enable extreme model compression through complementary parameter roles} 
\textbf{a}. Bit-budget landscape showing classification accuracy as a function of weight bits (y-axis) and delay bits (x-axis) for SHD, \textbf{b}. SSC, and \textbf{c}. NTIDIGITS. For each dataset, the left sub-panel reports the hidden-layer memory footprint, with delay memory shown as an offset corresponding to 0--7 delay bits, which contributes only a small fraction of the total. The right sub-panel shows accuracy across quantisation combinations (yellow indicates higher performance); dashed contours denote iso-performance boundaries. Black circles highlight two operating points: Solution~I, an accuracy-matched configuration competitive with strong baselines at markedly lower bit budgets, and Solution~II, which attains peak accuracy.
\textbf{d}. Parameter distribution histograms comparing high-efficiency   (Solution I, top panels) versus high-precision (Solution II, bottom panels) approaches for SHD, \textbf{e}. SSC, and \textbf{f}. NTIDIGITS dataset. Left panels show synaptic weight distributions and right panels show axonal delay distributions (time steps) across both first and second layers. {Across datasets and layers, learning reshapes the delay distributions under different weight quantisation levels.}
\textbf{g}. Memory-performance Pareto frontier. Each point denotes a quantisation configuration (x-axis: memory footprint, log scale; y-axis: classification accuracy). Introducing delay precision shifts the frontier outward, enabling accuracy-matched solutions at 20$\times$, 20$\times$, and 19$\times$ lower parameter-memory budgets for SHD (left), SSC (middle), and NTIDIGITS (right), respectively. Black dots indicate individual configurations, and the red dot marks the common solution (full-precision weights without delays).
}
\label{fig2}
\end{figure*}

\subsection{Exploiting delays reduces the need for weight precision, enabling high performance under extreme weight compression}

The finding that delays substantially improve performance in proportion to task temporal complexity prompts a natural follow-up question: Are precise weights necessary at all? Here, we focus on the memory-limited regime: rather than adding delays as an extra degree of freedom, we ask how temporal precision reallocates representational capacity when synaptic amplitudes are severely quantised. This question is particularly compelling given that biological synapses are inherently variable, highly noisy with unknown precision and subject to continual turnover \cite{rusakov2020noisy,oconnor2005graded,puro1977synapse}.

We hypothesized that weights and delays exhibit complementary sensitivities to distinct task statistics - specifically, that delays primarily capture temporal dependencies by shifting spike contributions in time. To test this weight-delay trade-off, we systematically quantized both parameter types independently across all tasks. Quantization reduces numerical precision, such as from 32-bit floating point (e.g., 3.14159265) to coarser representations like 2-bit (e.g., {-1, 0, 1, 2}) or even 1.58 bit (e.g., {-1, 0, 1}). Performance degradation under quantization reveals computational importance: stable performance indicates minimal reliance on precise values. Notably, this directly informs neuromorphic hardware deployment because processors operate under strict ``bit-budgets" that constrain parameter allocation \cite{davies2021advancing,furber2014spinnaker,schuman2022opportunities,dampfhoffer2023backpropagation, kudithipudi2025neuromorphic}.

\textbf{Figure \ref{fig2}a-c} shows model performance across the ``bit-budget" trade-off landscape between weights and delays for SHD, SSC and NTIDIGITS, respectively. Strikingly, for these temporally complex tasks, we found that the introduction of minimal delay bits provides significant improvement in performance commensurate with full-precision results. {Here, {total bits} denote the sum of the weight-bit and delay-bit allocations used in our bit-budget protocol (see \textbf{Methods \ref{bits}}).} For example, we identified highly-efficient solutions (solution I) that match full precision baseline performance while using only 4.58, 7.58 and 7.58 total bits, compared with 35, 35 and 36 bits for SHD, SSC, and NTIDIGITS respectively. This is an approximate 7$\times$, 4$\times$ and 4$\times$ reduction respectively. Moreover, our quantization analysis makes it possible to investigate how, for the same number of bits, one can still have much more performant solutions by combining mixtures of precision values of delay versus weight parameters \cite{grappolini2023beyond,hazan2022memorytemporaldelaysweightless}. For instance, on SHD, 4-bit weights with 5-bit delays yield 94.04\% accuracy (peak performance, solution II), matching state-of-the-art synaptic delay-enabled networks \citep{hammouamri2024learning,sun2025towards} demonstrating that delays enable comparable performance with only 28.56\% of the parameters and 4\% of the memory footprint. We find, in most instances, using only 1.58-bit precision (with weights being only negative, absent or positive) was sufficient to recover accuracy comparable to strong full-precision baselines when combined with learned delays. When weights were trained alone without delays, performance remained relatively poor across weight quantization levels (\textbf{Supplementary Figure \ref{sup3}}).

Since bits represent the fundamental unit of digital memory, the weight quantization we apply translates directly to storage requirements which we show to the left of the ``bit budget" landscapes (\textbf{Figure \ref{fig2}a, b, c}). We illustrate in \textbf{Figure \ref{fig2}d, e, f} the learned weight and delay distributions of highlighted solutions I (top) and II (bottom) in the ``bit budget" landscape, for each task. In \textbf{Figure \ref{fig2}g} we summarize the memory-performance trade-offs across task, revealing that delay-based networks can be accuracy-matched to strong full-precision baselines while reducing parameter memory by 20$\times$, 20$\times$, 19$\times$ on SHD, SSC, NTIDIGITS respectively. \textbf{Supplementary Figure \ref{sup4}} shows how these results scale with the number of neurons and in \textbf{Supplementary Figure \ref{sup5}} we provide a detailed analysis of the relative contribution of the delays to performance. {\textbf{Supplementary Figure~\ref{weightdelaycodesign}} shows that random heterogeneous delays (uniform and Gaussian) could improve performance under extreme weight quantisation, but trainable delays yield the largest gains by learning a task-matched delay distribution.
} {As a further stress test of the bit-budget setting, we constrained each layer to two weight values: a positive value and zero. Accuracy deteriorated under this restriction, indicating that allowing inhibitory (negative) weights is necessary at least in the readout layer (\textbf{Supplementary Table~{\ref{excitatory}}}).}

Together, these findings indicate that introducing axonal delays enables networks to tolerate dramatically lower weight precision, maintaining accuracy matched to strong full-precision baselines with approximately 20x fewer weight bits. This extreme memory efficiency stems from delays' ability to capture temporal dependencies that cannot be easily encoded through weight precision alone.

\begin{figure*}[]
\centering
\includegraphics[width=1\linewidth]{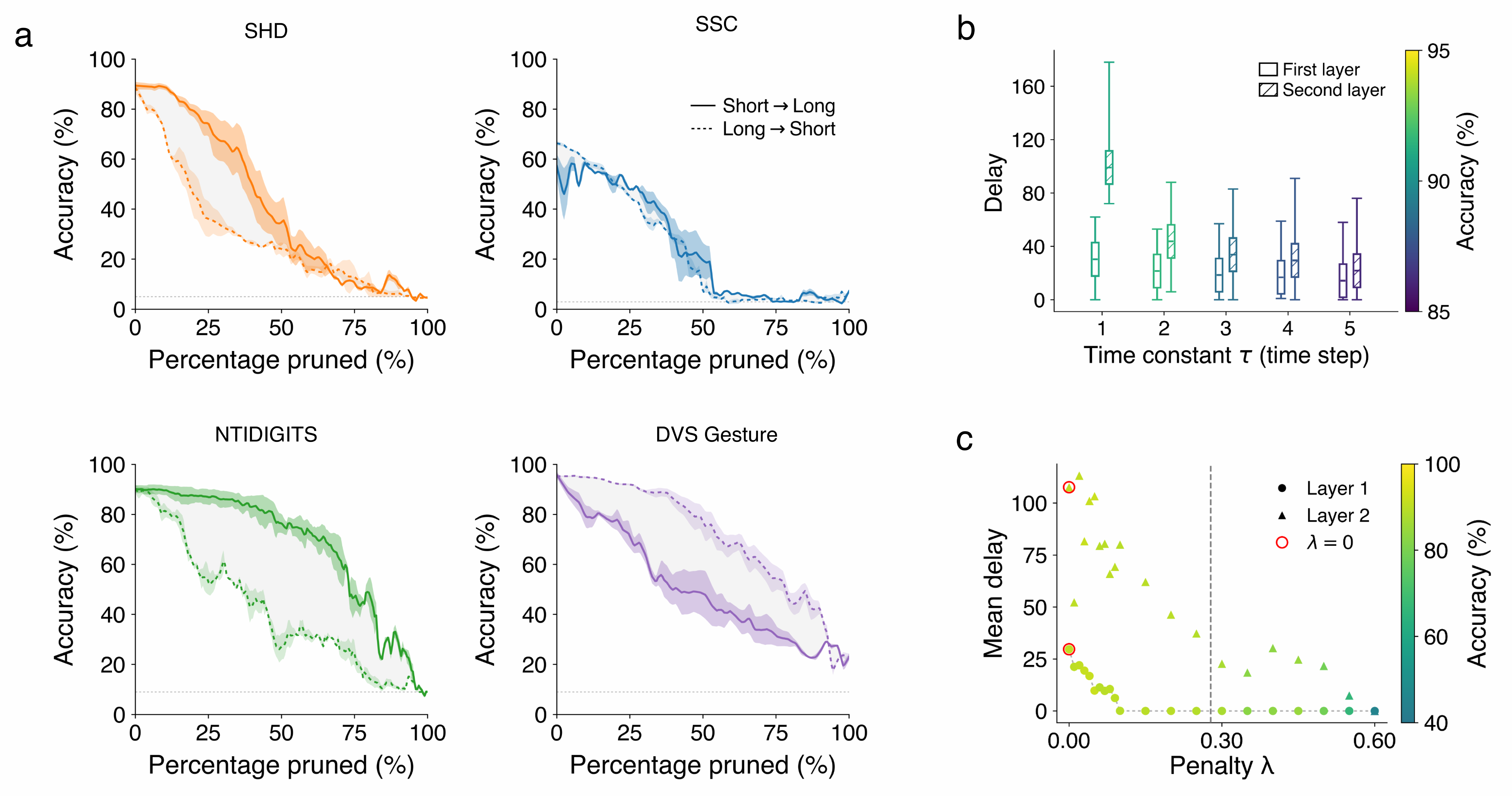}
\caption{{
Performance is selectively sensitive to long delays and can be partially traded off with time constants and regularisation.}
\textbf{a}. Weight connection-pruning ablation on SHD, SSC,  NTIDIGITS and DVS Gesture. Test accuracy is plotted against the fraction of delays removed either \emph{Short$\rightarrow$Long} (dark solid line) or \emph{Long$\rightarrow$Short} (light dashed line). Performance deviation represents standard error across n = 5 independent training runs.  Pruning many neurons with short delays has a relatively little effect, whereas removing a small set of the longest delays causes a sharp drop in accuracy for the well-controlled datasets (SHD and NTIDIGITS). DVS Gesture shows minimal sensitivity for long delays, consistent with the lower temporal complexity of the task. \textbf{b}. Time constant trade-off. With plastic delays, increasing the layer-shared time constant ($\tau$) systematically shifts the learned delay distributions towards shorter values and reduces dispersion. A non-trivial long tail persists, especially in deeper layers, indicating that longer $\tau$ only partially substitutes for long-range delays. Accuracy decreases at excessively large $\tau$. \textbf{c}. Relationship between delay and $L_{2}$ regularisation (strength $\lambda$). As $\lambda$ increases, the mean delay magnitude and long-delay fraction decrease while test accuracy remains high. The vertical dashed line marks 89\% accuracy (higher to the left). Earlier-layer delays contract first, whereas long delays in later layers are preferentially preserved. The horizontal dashed lines correspond to the zero-delay network. }
\label{fig3}
\end{figure*}

\subsection{The relative importance of delay lengths is task-dependent and trades off with neural time constants}

Having established that training delays enhances performance significantly beyond weights alone, we next asked \textit{which} delays are critical: are all delays equally important, or does the relative importance of delay lengths depend on the task? To answer this, we conducted systematic ablation studies, sequentially removing synaptic connections ranked by axonal delay magnitude from shortest to longest and vice versa.

Our findings revealed striking task-dependent asymmetry (\textbf{Figure~\ref{fig3}a}). For well-controlled temporally complex tasks (SHD and NTIDIGITS), pruning connections with longer delays caused dramatically steeper performance decline compared to removing short delays. DVS Gesture showed the opposite pattern: performance degraded more steeply when short delays were ablated, consistent with this task requiring temporal sensitivity to the shorter timescales intrinsic to motion {(consistent with fast motion detection via delays \citep{grimaldi2023learning})}. In \textbf{Supplementary~Figure~\ref{sup6}} we further show that directly ablating long delays has a significantly greater impact on performance than ablating short delays on the auditory tasks. These ablation results demonstrate that networks attune their delay distributions to task-intrinsic timescales rather than universally prioritizing any particular delay range. {\textbf{Supplementary Figure~\ref{supdelaymax}} extends this analysis by imposing a maximum-delay constraint during training. Performance increases with the allowable maximum delay and then saturates on the SHD dataset, consistent with a need for long-range temporal dependencies.
}

We next asked whether the magnitude of delays could be reduced while preserving the computational benefits that their heterogeneity provides. This question is motivated by both practical and biological considerations. Practically, learned delays introduce latency and buffering costs \citep{meijer2025efficient} that are undesirable in resource-limited settings. Biologically, neural systems may employ complementary mechanisms to reduce reliance on potentially costly long-range delays while maintaining temporal computation. We investigated two complementary mechanisms: (1) neural time constants \cite{perez2021neural} and (2) explicit regularization \cite{achterberg2023spatially}.

First, we examined the relationship with neural time constants ($\tau$), another key temporal parameter to the delay likely important for learning in changing environments \cite{perez2021neural}. In biological networks, time constants are typically shorter than axonal delays, enabling the physical chain of receiving, integrating, spiking, and transmitting that underlies spatiotemporal computation \cite{keller2024spacetimeperspectivedynamicalcomputation}. We hypothesized longer time constants could partially compensate for delays by maintaining information for longer. Systematically varying $\tau$ while keeping delays plastic revealed that delay distributions shifted systematically: as time constants increased, learned delays shortened and long tails contracted, though non-trivial long-tails persisted even at longest time constants, particularly in deeper layers (\textbf{Figure \ref{fig3}b}). The learned delay distributions can be found in \textbf{Supplementary Figure \ref{sup7}} which shows clearly how delays shift in accordance to $\tau$.

\begin{figure*}[]
\centering
\includegraphics[width=1\linewidth]{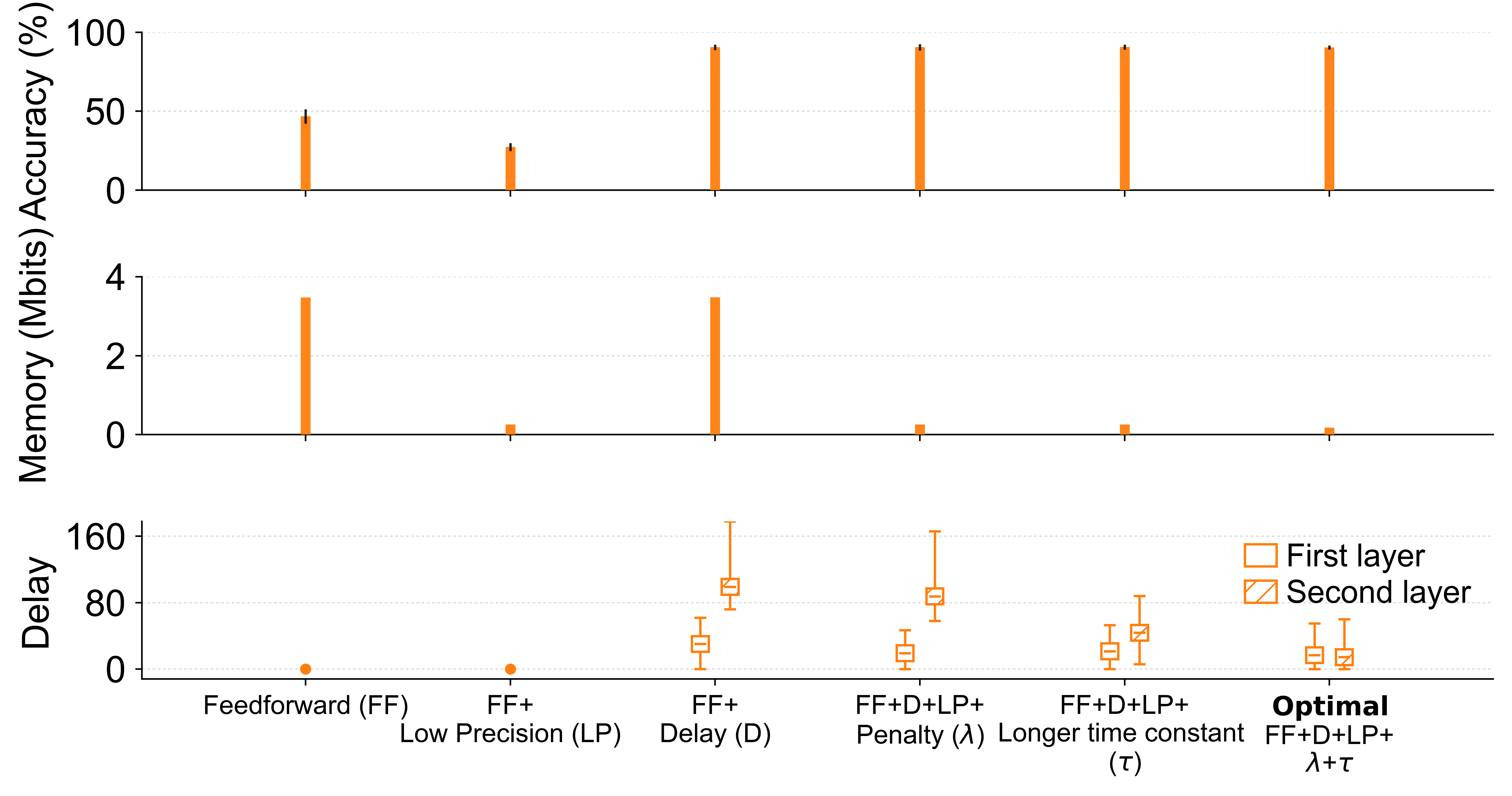}
\caption{\textbf{Delay learning, quantization, regularization, and time constants jointly optimize resource efficiency on SHD dataset.} (Top) Test accuracy (\%) with error bars representing 3$\sigma$; delay-enabled configurations achieve similar accuracies. (Middle) Memory footprint (Mbits) demonstrates parameter-efficient gains from delay heterogeneity, enabling compact models without sacrificing accuracy. (Bottom) Learned axonal delay distributions per layer under distinct configurations. Starting from weight-only feedforward baselines, direct weight quantization reduces memory but degrades accuracy. Adding learned delays with regularization and longer time constants ($\tau$) restores high accuracy while achieving substantially lower memory footprint. The optimal solution combines 1.58-bit weight quantization, 5-step delay quantization, delay regularization, and $\tau=2$, achieving 90\% accuracy with only 0.174 Mbits and sparse long-delay distributions.}
\label{fig4}
\end{figure*}

Second, we imposed direct $L_2$ regularization (strength denoted by $\lambda$) on delay magnitudes. We found that a small amount of regularization increased accuracy while simultaneously decreasing mean delays (\textbf{Figure \ref{fig3}c}). At $\lambda>0.10$, first hidden layer delays collapsed toward zero - indicating a hierarchical preference for decreasing delays in the earlier versus later layers. Increasing $\lambda$ produced monotonically declining delays, initially rising before falling, yet consistently exceeding baseline. Interestingly, test accuracy remained above 89\% throughout, confirming that regularization selectively eliminates redundant short delays (especially in early layers) while preserving critical long-delay connections.

Together, these results demonstrate that computational efficiency emerges from delay distributions matched to task requirements. For well controlled temporally complex tasks, this means sparse long delays working in concert with time constants which can partially absorb the temporal span. Regularization reveals task-dependent hierarchical organization in which early layers can operate with minimal delays while later layers require longer delays.

\subsection{A highly compact neuromorphic architecture via joint parameter optimization}

Having established these effects across analyses, we synthesized a final architecture that combines low-bit weight quantisation with learnable, regularised axonal delays. This exploits the complementary strengths of delays and weights: delays provide parameter-efficient temporal alignment, compensate for aggressive weight quantization to maintain performance despite extreme compression, and can be selectively retained (favoring sparse long delays) through time constant trade-offs and explicit regularization constraints.

We systematically evaluated the incremental effect of each design constraint on performance, memory, and delay characteristics (\textbf{Figure \ref{fig4}}), culminating in the final optimized architecture. This solution combines regularization on delays, 1.58-bit weight quantization, delay quantization with 5-step resolution, and fixed time constant ($\tau$) at two time steps. This configuration attained 90\% accuracy on SHD with only 0.174 Mbits total memory. With 50\% weight sparsity, learned delays exhibited mean 16.9 steps (max 55) in the first hidden layer and mean 14.6 steps (max 60) in the second. The resulting architecture retains only essential delays while minimizing memory footprint, demonstrating that delay heterogeneity can be systematically optimized as a tunable resource within physical constraints.

\section{Discussion}\label{sec12}
{Our central contribution is to quantify when temporal heterogeneity can substitute for synaptic precision under fixed parameter-memory budgets, reframing delays as an allocatable representational resource in time.} We demonstrate that introducing even a small amount of learnable delay heterogeneity can provide computational benefits beyond simply learning synaptic weights and enable highly performant solutions even when weights are aggressively compressed, as we expect in nature \cite{shuvaev2024encoding}. This is significant because it reveals that temporal information can support computation without requiring high-precision weights, given the substantial costs of memory access and data transfer \cite{wu2026dendritic} and the increasing use of weight quantization at inference for large-scale AI deployment \cite{lang2024comprehensivestudyquantizationtechniques}, while resonating with the known imprecision of synaptic weights in biological brains \cite{rusakov2020noisy,ma2023exploiting}.

These findings open several directions for future research. While we focused on relatively simple architectures to isolate the role of delays, to fully assess generality future work should explore how delay-based principles can be incorporated into larger transformer or hybrid architectures \cite{Augustin_2025, zhu2025hybrid}, particularly to advance promising spiking-based transformer architectures that currently still only learn weights \cite{guo2025spikingtransformerintroducingaccurateadditiononly}. Fine-tuning delays in neuromorphic spiking networks initialized from pre-trained weighted baselines may be a particularly interesting approach (as we show is possible \textbf{Supplementary Figure \ref{sup2}}). Beyond strictly neuromorphic settings, these principles may extend to asynchronous distributed inference systems, from accelerator coordination systems \cite{barham2022pathways} to distributed multi-agent robotics \cite{murai2024robot} or always-on event-based vision \citep{Grimaldi24}, where heterogeneous communication delays naturally emerge and could be strategically exploited rather than minimized.

A key insight from our work is that spikes may derive much of their computational power from their natural capacity to represent temporal information. While we tested our approach on the most widely available neuromorphic benchmarks, these datasets remain limited in capturing the full complexity of real-world temporal dynamics. Developing and exploring new benchmark tasks that are truly embodied, testing these principles in dynamic environmental contexts operating in physical space, would better reflect real-world scenarios \cite{yik2025neurobench}.

Our focus on axonal delays was motivated by computational tractability, as they scale linearly with neuron count ($O(n)$). The rationale was that if axonal delays alone yield substantial benefits, delays with greater parameter complexity, such as dendritic and synaptic delays as found in the brain \cite{agmonsnir1993signal,swadlow1987corticogeniculate}, should provide even richer computational advantages at multiple timescales. Our purpose was not to fully replicate the brain's empirical delay distribution but rather to demonstrate how delay heterogeneity can be leveraged. Incorporating synaptic and dendritic delays represents a natural next step, both for examining how distinct heterogeneous delay types further augment our findings and for enabling direct comparisons in bit allocations for identical numbers of parameters in weights versus delays (e.g., $n^2$ synaptic weights versus $n^2$ synaptic or dendritic delays).

Finally, we see immense promise in extending this work to explore additional forms of heterogeneity that incorporate spatial embedding \cite{achterberg2023spatially,lundqvist2023working,keller2024spacetimeperspectivedynamicalcomputation}. Such extensions would more directly link computation through both space and time, deepening our understanding of novel computational principles spanning both embodied biological and artificial intelligent systems. We also envisage extending this perspective to embodied intelligence, exploring how learning under physical and resource constraints can exploit spatial and temporal heterogeneity to support interactions with dynamic environments \citep{yu2025selfmotionstructuralpriorcoherent}.

\section{Methods}\label{sec:methods}

\subsection*{Spiking neuron model and axonal delays}

Neuronal dynamics were simulated using the Spike Response Model (SRM) \citep{gerstner1995time}.  
For neuron~$i$ in layer~$l$, the membrane potential and spike emission are defined as
\begin{gather}
  u_i^l(t) = \sum_j W_{ij}^{l-1}\bigl(\varepsilon * s_j^{l-1}\bigr)(t)
             + \bigl(\nu * s_i^l\bigr)(t), \\
  s_i^l(t) = \Theta\bigl(u_i^l(t) - \theta_u\bigr),
\end{gather}
where $W_{ij}^{l-1}$ denotes the synaptic weight from presynaptic neuron $j$ to postsynaptic neuron $i$,  
$s_j^{l-1}(t)$ is the presynaptic spike train, and $\theta_u$ is the membrane potential threshold.

The spike response kernel was defined as 
$\varepsilon(t) = \tfrac{t}{\tau_s}\exp\left(1 - \tfrac{t}{\tau_s}\right)\Theta(t)$, 
and the refractory kernel as 
$\nu(t) = -2\theta_u \tfrac{t}{\tau_r}\exp\left(1 - \tfrac{t}{\tau_r}\right)\Theta(t)$, 
with synaptic time constant $\tau_s$ and refractory time constant $\tau_r$. 
The Heaviside step function $\Theta(\cdot)$ outputs 1 if its argument is positive and 0 otherwise.

Axonal transmission delays were modeled by shifting each neuron’s output spike train by a learnable delay $d_i^l$:
\begin{equation}
  s_{d,i}^l(t) = \bigl(\delta(t - d_i^l) * s_i^l\bigr)(t),
  \quad d_i^l \in [0, \theta_d],
\end{equation}
where $\delta(\cdot)$ is the Dirac delta function, and $\theta_d$ specifies the maximum allowable delay.  
All outgoing synapses from neuron $i$ in layer $l$ share the same axonal delay $d_i^l$.  

The synaptic weights were optimized using surrogate-gradient Backpropagation Through Time (BPTT) \citep{neftci2019surrogate}.  
The surrogate gradient of the firing function was defined as
\begin{equation}
  \hat{f}_s'(u) = \tau_{\mathrm{scale}}
  \exp\left(-\frac{\lvert u-\theta_u \rvert}{\tau_\theta}\right)
\end{equation}
where $\tau_{\mathrm{scale}}$ controls the gradient magnitude and $\tau_\theta$ sets the sharpness around the threshold $\theta_u$ \citep{shrestha2018slayer}.  

The gradient of the loss $E$ with respect to a delay parameter $d^l$ is
\begin{equation}
  \nabla_{d^l} E = T_s \sum_{n=0}^{N_s} \frac{\partial \mathcal L[n]}{\partial d^l},
\end{equation}
where $T_s$ is the simulation step size, $N_s$ is the number of time steps, and $\mathcal L[n]$ is the loss contribution at step $n$.  
Expanding with the chain rule,
\begin{equation}
\begin{split}
  \nabla_{d^l} E
  &= T_s \sum_{n=0}^{N_s} \sum_{m=0}^{n} 
     \frac{\partial s_d^l[m]}{\partial d^l} 
     \frac{\partial   \mathcal L[n]}{\partial s_d^l[m]} \\
  &\approx - T_s \sum_{n=0}^{N_s} \sum_{m=0}^{n} 
     \frac{s_d^l[m] - s_d^l[m-1]}{T_s} 
     \frac{\partial  \mathcal L[n]}{\partial s_d^l[m]},
\end{split}
\end{equation}
where $\partial s_d^l[m]/\partial d^l$ is numerically approximated using a finite difference $-(s_d^l[m]-s_d^l[m-1])/T_s$ \citep{shrestha2018slayer,sun2023learnable}.

\subsection*{Synaptic weight and axonal delay quantization}
\label{bits}

We adopted a quantization scheme inspired by the DoReFa-Net framework \citep{zhou2016dorefa}.
In the general $m$-bit case, a real-valued weight $W_{ij}\in[0,1]$ was first mapped to an integer representation,
\[
  W_o = \mathrm{round}\bigl((2^m - 1)W_{ij}\bigr), \quad W_o \in [0,2^m-1],
\]
and then rescaled to obtain a quantised value $W_q \in [-1,1]$,
\[
  W_q = \frac{2W_o}{2^m-1} - 1.
\]

Beyond uniform $m$-bit quantization, we also examined extreme low-bit settings.
For ternary quantization (corresponding to $\log_2 3 \approx 1.58$ effective bits), a threshold was introduced,
\[
  \delta_w = 0.7\,\langle |W_{ij}| \rangle,
\]
where $\langle |W_{ij}| \rangle$ denotes the average absolute weight magnitude.
Weights were then discretized as
\[
  W_q =
  \begin{cases}
    \alpha, & W_{ij}>\delta_w, \\
    0,      & |W_{ij}|\le \delta_w, \\
    -\beta, & W_{ij}<-\delta_w,
  \end{cases}
\]
with $\alpha,\beta>0$ as scaling factors, either fixed ((e.g. $\alpha=\beta=1$)) or learnable. 
When learnable, the gradients propagated according to
\[
  \frac{\partial \mathcal{L}}{\partial \alpha}
  = \sum_{W_{ij}>\delta_w} \frac{\partial \mathcal{L}}{\partial W_q},
  \qquad
  \frac{\partial \mathcal{L}}{\partial \beta}
  = - \sum_{W_{ij}<-\delta_w} \frac{\partial \mathcal{L}}{\partial W_q}.
\]

Axonal delays were quantized onto a uniform grid with lower-bound offset $\delta_d$ and step size $\Delta_d>0$. 
Given a raw (continuous integer) delay $d$, the forward mapping applied a round-down (mid-rise) uniform quantizer:
\begin{equation}
  \hat{d} =
  \biggl\lfloor \frac{d - \delta_d}{\Delta_d} \biggr\rfloor \Delta_d + \delta_d,
\end{equation}
where $\delta_d$ specifies the grid offset (interpreted as the minimal resolvable delay) and $\Delta_d$ is the quantisation step. 
This operation maps $d$ to the greatest grid point not exceeding $d$ on the lattice $\{\delta_d + k\Delta_d \mid k\in\mathbb{Z}\}$.

In backpropagation, a straight-through estimator (STE) \citep{bengio2013estimating}  passed gradients unchanged:
\begin{align}
  \frac{\partial L}{\partial d} &= \frac{\partial L}{\partial \hat{ d}}, \\
  \frac{\partial L}{\partial \delta_d} &= \frac{\partial L}{\partial \hat{ d}}.
\end{align}

\subsection*{Spikemax classification loss}

\begin{table}[]
\centering
\caption{\textbf{Notation summary.} Key variables used in the spiking neuron model, quantization, and loss functions.}
\label{tab:notation}
\begin{tabular}{ll}
\toprule
\textbf{Symbol} & \textbf{Description} \\
\midrule
$u_i^l(t)$ & Membrane potential of neuron $i$ in layer $l$ at time $t$ \\
$s_i^l(t)$ & Spike output of neuron\\
$W_{ij}^{\,l-1}$ & Synaptic weight\\
$\theta_u$ & Membrane potential threshold \\
$\tau_s$, $\tau_r$ & Synaptic and refractory time constants \\
$\tau_{\mathrm{scale}}, \tau_\theta$ &  scaling factor and sharpness around $\theta_u$ \\
$\varepsilon(t)$, $\nu(t)$ & Synaptic response kernel; refractory kernel \\
$d_i^l$, $\theta_d$ & Learnable axonal delay of neuron; maximum allowable delay \\
$T_s$, $N_s$ & Simulation step size; total number of simulation steps \\
\midrule
$m$ & Bit-width for quantization \\
$\delta_w$, $\alpha,\beta$ & Weight quantization threshold; scaling factors \\
$\delta_d$, $\Delta_d$ & Minimum  delay; quantization step size for delays \\
\midrule
$c_i(t)$, $p_i(t)$ & Spike count in readout window; instantaneous class probability \\
$T$, $W$, $n_l$ & Total input duration; spike-counting window; index of readout layer \\
$\mathcal L$, $\mathcal L_{\text{reg}}$, $\lambda_l$ & Training loss; $L_2$ delay regularization; layer-specific penalty coefficient \\
$f_d$ & Delay-bit fraction used in bit-budget analysis \\
\bottomrule
\end{tabular}
\end{table}

\begin{table}[]
    \centering
    \caption{\textbf{Hyperparameter settings across datasets.} 
  }
    \label{parameter_hyper}
    \begin{tabular}{lccccccc}
        \toprule
        \textbf{Dataset}  & $\tau_s$ & $\tau_r$ & $\tau_{\mathrm{scale}}, \tau_\theta$ & $\theta_u$ & $\Delta_d$ & $\delta_d$ & $\theta_d$\\
        \midrule
        SHD            & 1 & 2 & (0.1, 1)   & 10 & $[1, 70]$ & 0/learnable &$+\infty$\\
        SSC           & 1 & 1 & (0.1, 1)  & 10  & $[1, 70]$& 0/learnable& $+\infty$\\
        NTIDIGITS      & 1 & 1 & (0.1, 1)   & 10 &$[1, 70]$& 0/learnable &$+\infty$\\
        DVS Gesture  & 5 & 5 & (0.1, 1) & 10 & $[1, 70]$ & 0/learnable&$+\infty$\\
        \bottomrule
    \end{tabular}
\end{table}

\begin{table*}[t]
    \centering
    \caption{\textbf{Network configurations across datasets.} {Input size refers to cochleagram channels (SHD/SSC), auditory features (NTIDIGITS), or event-camera resolution (DVS Gesture). For DVS Gesture, we use two spiking convolutional layers (16 and 32 channels) followed by one spiking fully connected layer (512 neurons).}}
    \label{tbl:networks}
    \begin{tabular}{lcccc}
        \toprule
        \textbf{Dataset} & \textbf{SHD} & \textbf{SSC} & \textbf{NTIDIGITS} & \textbf{DVS Gesture} \\
        \midrule
        Input size & 700 & 700 & 64 & 128$\times$128$\times$2 \\
        Number of hidden layers & 2 & 2 & 2 & 3 \\
        Hidden spiking units & 128 & 128 & 256 & - \\
        Output spiking neurons & 20 & 35 & 11 & 11 \\
        \bottomrule
    \end{tabular}
\end{table*}

We used the Spikemax loss \citep{shrestha2022spikemax}, which derives class probabilities from maximal spike counts in the readout layer.  
Spike counts $c_i(t)$ for neuron $i$ were aggregated over a sliding window of length $W$ (typically equal to the full sequence length, $W=T$):
\begin{equation}
  c_i(t) = \int_{t-W}^t s_i^{(n_l)}(\tau)\,\mathrm d\tau, 
  \qquad
  p_i(t) = \frac{c_i(t)}{\sum_j c_j(t)},
\end{equation}
where $s_i^{(n_l)}(\tau)$ is the spike train of readout neuron $i$ in the final layer $n_l$, and $p_i(t)$ denotes the instantaneous class probability.  

The overall loss is given by
\begin{equation}
  \mathcal L = \frac{1}{T} \int_0^T \Bigl(-\sum_i \hat y_i \ln p_i(t)\Bigr)\,\mathrm d t,
\end{equation}

where $\hat y_i$ is the one-hot target label for class $i$ and $T$ is the total input duration.  This loss encourages early accumulation of class evidence rather than relying solely on late spikes.

\subsection*{Delay regularization}

To discourage excessively long delays, we applied an $L_2$ penalty:
\begin{equation}
  \mathcal L_{\text{reg}} = \sum_l \lambda_l \sum_i (d_i^l)^2,
\end{equation}
where $d_i^l$ is the axonal delay of neuron $i$ in layer $l$, and $\lambda_l$ is a layer-specific regularization coefficient.  

Synaptic weights were initialised with Kaiming normal initialisation. Axonal delays were initialised i.i.d.\ as $d_i^l \sim \mathcal{U}(0,1)$ (in time steps).
Unless otherwise stated, neurons within the same layer shared identical membrane thresholds $\theta_u$, kernel parameters, quantization bit-widths, and minimum delay thresholds. A summary of notation and hyperparameters is given in Tables~\ref{tab:notation},~\ref{parameter_hyper} and~\ref{tbl:networks}. 
All experiments were implemented in the SLAYER framework \citep{shrestha2018slayer} on an NVIDIA GeForce GTX 1080Ti GPU.

\bmhead{Acknowledgements}
This project was funded by the Advanced Research + Invention Agency (ARIA). D.A. is funded by an Imperial College Research Fellowship, Schmidt Sciences Fellowship and Templeton World Charity Foundation, Inc (funder DOI 501100011730) under the grant TWCF-2022-30510. For the purpose of open access, the authors have applied a Creative Commons Attribution (CC BY) license to the text, figures and code relating to this paper.

\begin{appendices}

\end{appendices}

\bibliography{sn-bibliography}

\clearpage
\appendix

\section*{Supplementary Information}
\section*{Supplementary Figures}
\setcounter{suppfigcounter}{0}
\setcounter{figure}{0}
\renewcommand{\thefigure}{S\arabic{figure}}
\setcounter{table}{0}
\renewcommand{\thetable}{S\arabic{table}}

\suppfigsection{ Synaptic weight
and axonal delay distributions over the course of training for tasks }
{sup1}
\begin{figure}[H]
\centering
\includegraphics[width=\linewidth]{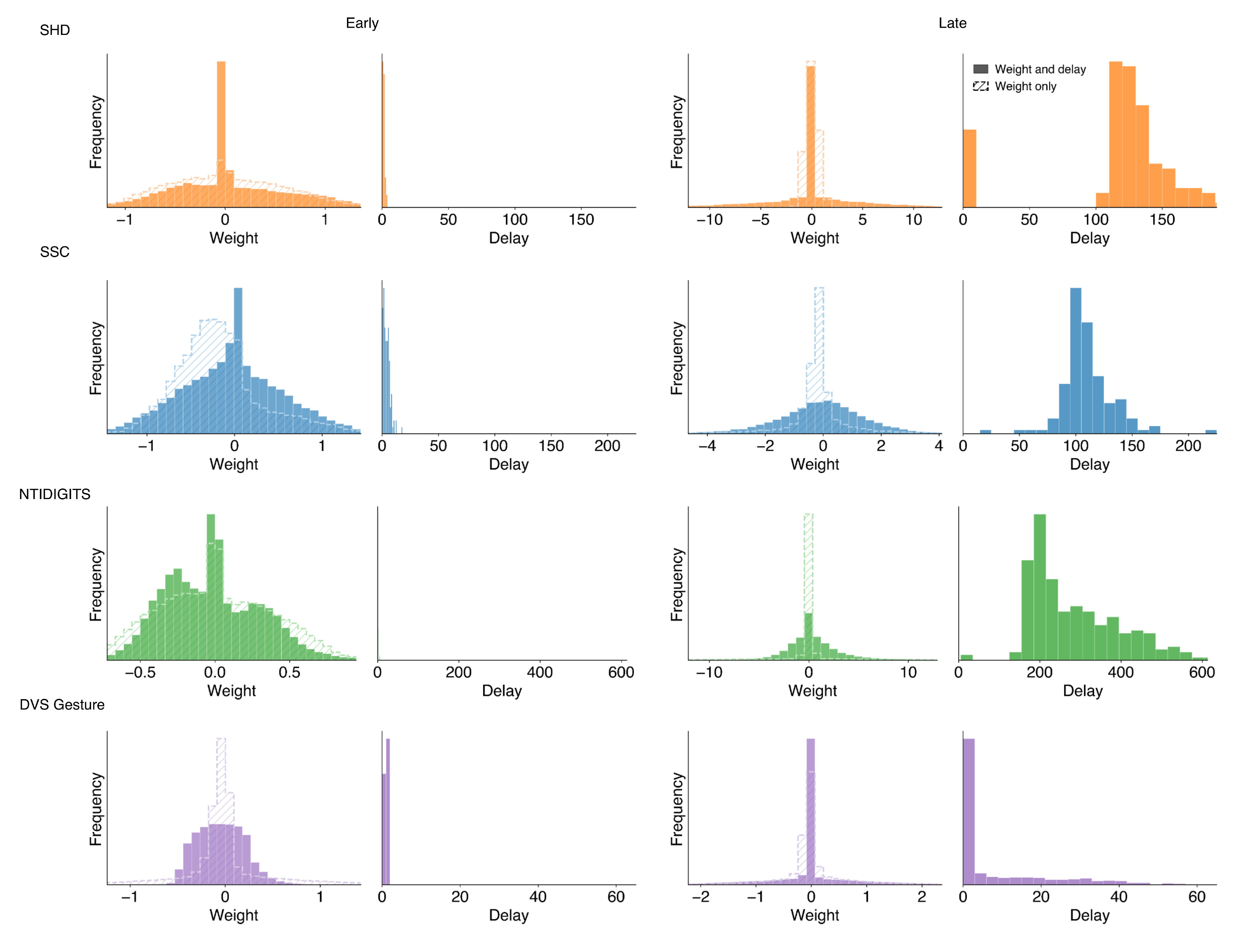}
\label{sup1_figure}
\end{figure}

The left panels show early-stage weight and delay distributions (Kaiming-initialised weights; delays sampled in [0,1] time steps), whereas the right panels show the corresponding distributions after training.

\clearpage

\suppfigsection{ Delays enable reservoir-like computation with frozen weights}{sup2}
\begin{figure}[H]
\centering
\includegraphics[width=0.45\linewidth]{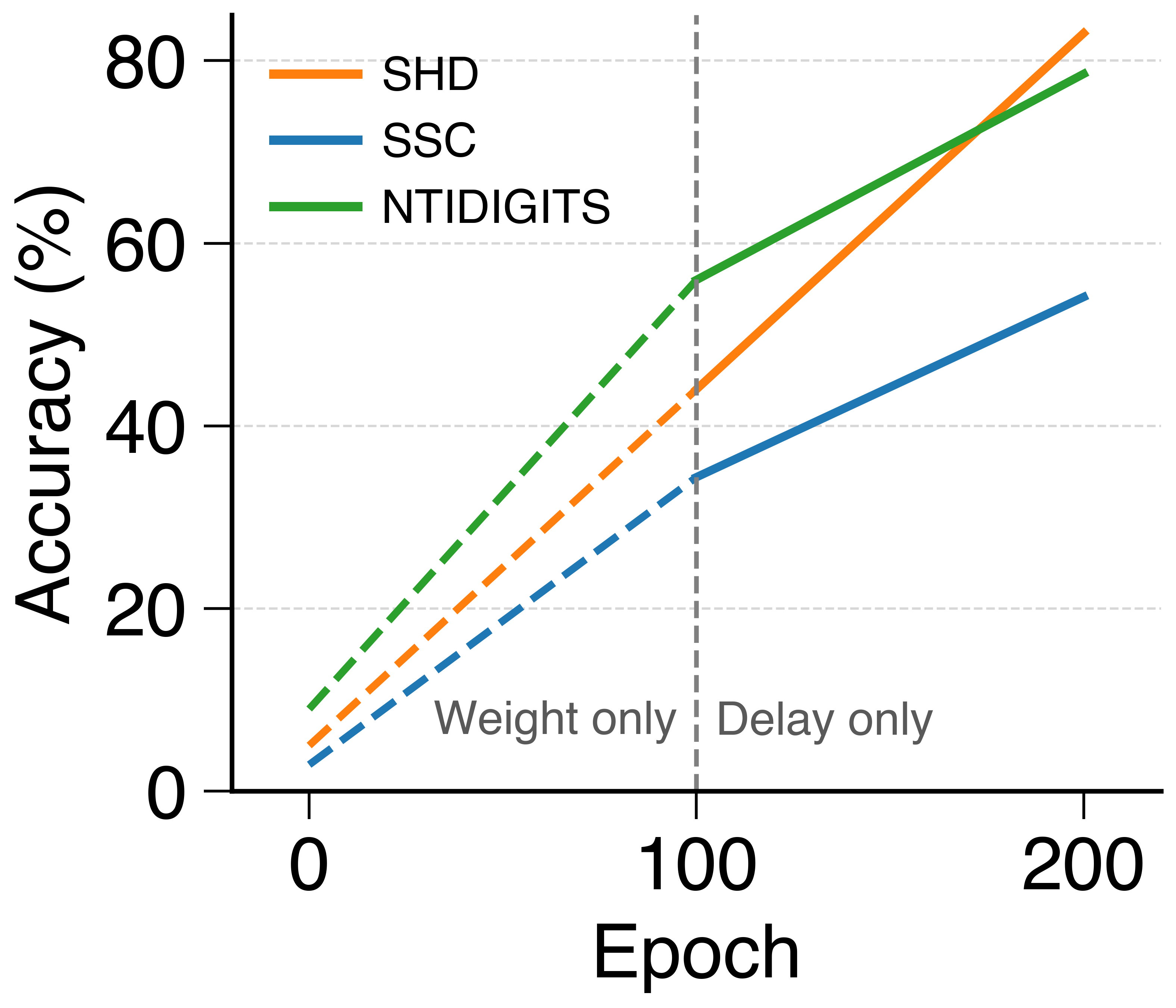}
\includegraphics[width=0.45\linewidth]{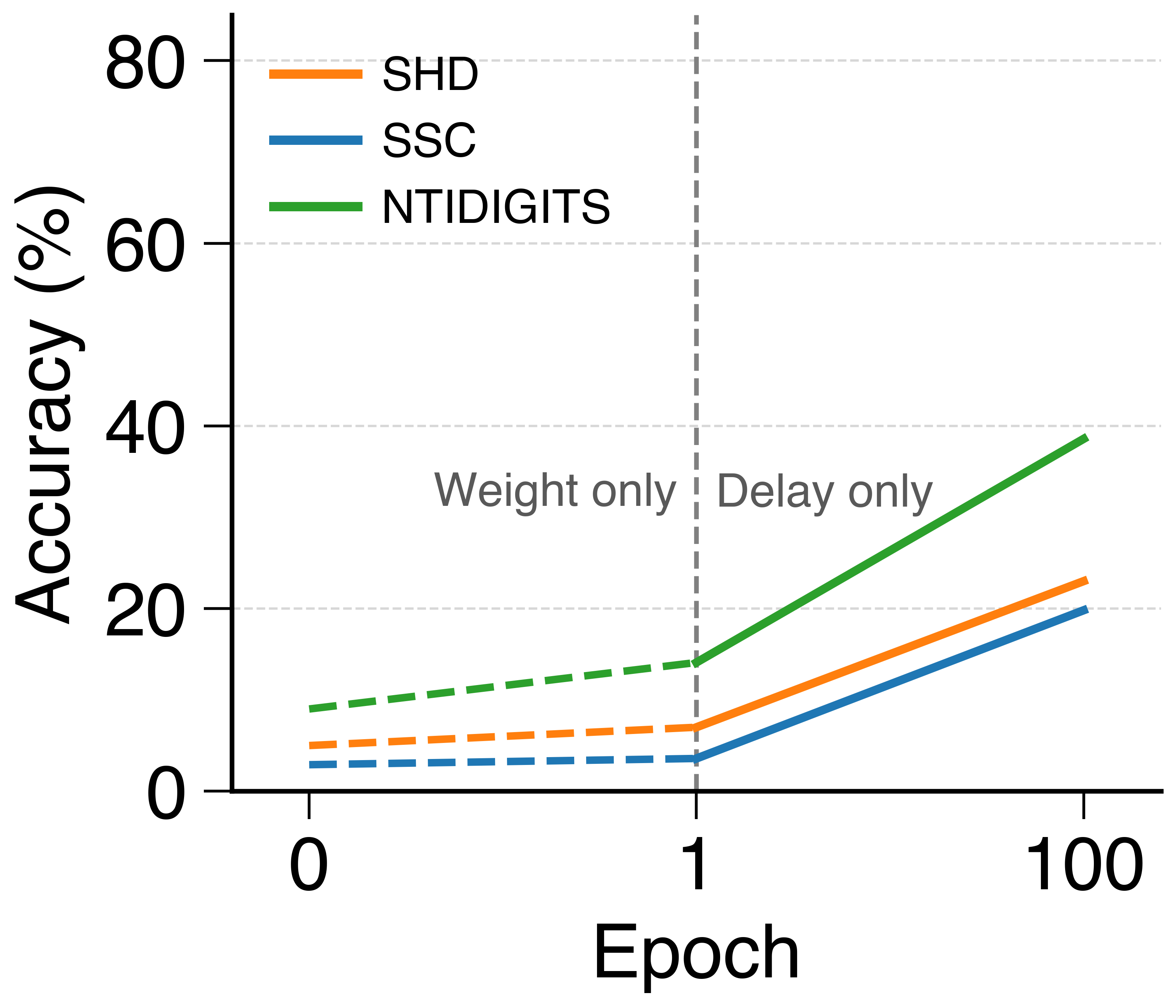}
\label{sup2_figure}
\end{figure}
To assess whether delays could compensate for fixed connectivity, we adopted a reservoir-like approach: feedforward SNNs with weight-only synapses were pretrained for varying numbers of epochs, after which weights were frozen, delays were introduced, and only axonal delays were optimised. Performance peaked with 100 pretraining epochs (left panel), after which delay-only optimisation improved accuracy by 39\% on SHD, 19\% on SSC, and 23\% on NTIDIGITS. Even with minimal pretraining (single epoch), accuracy increased by over 18\% across all datasets (right panel). These findings indicate that once stable dynamics are established, tuning axonal delays alone yields substantial gains on temporally complex tasks, highlighting delay learning as a computational paradigm for temporal structure exploitation. {Moreover, unfreezing the weights after delay introduction and continuing joint training recovers performance comparable to training weights and delays jointly from scratch. This supports a practical schedule in which, when weight-only networks are insufficient on temporally structured tasks, delays are introduced as a fine-tuning stage after weight dynamics have stabilised.}

\clearpage
\suppfigsection{ Weight precision analysis reveals task-dependent requirements}{sup3}
\label{weightprecision}
\begin{figure}[H]
\centering
\includegraphics[width=0.75\linewidth]{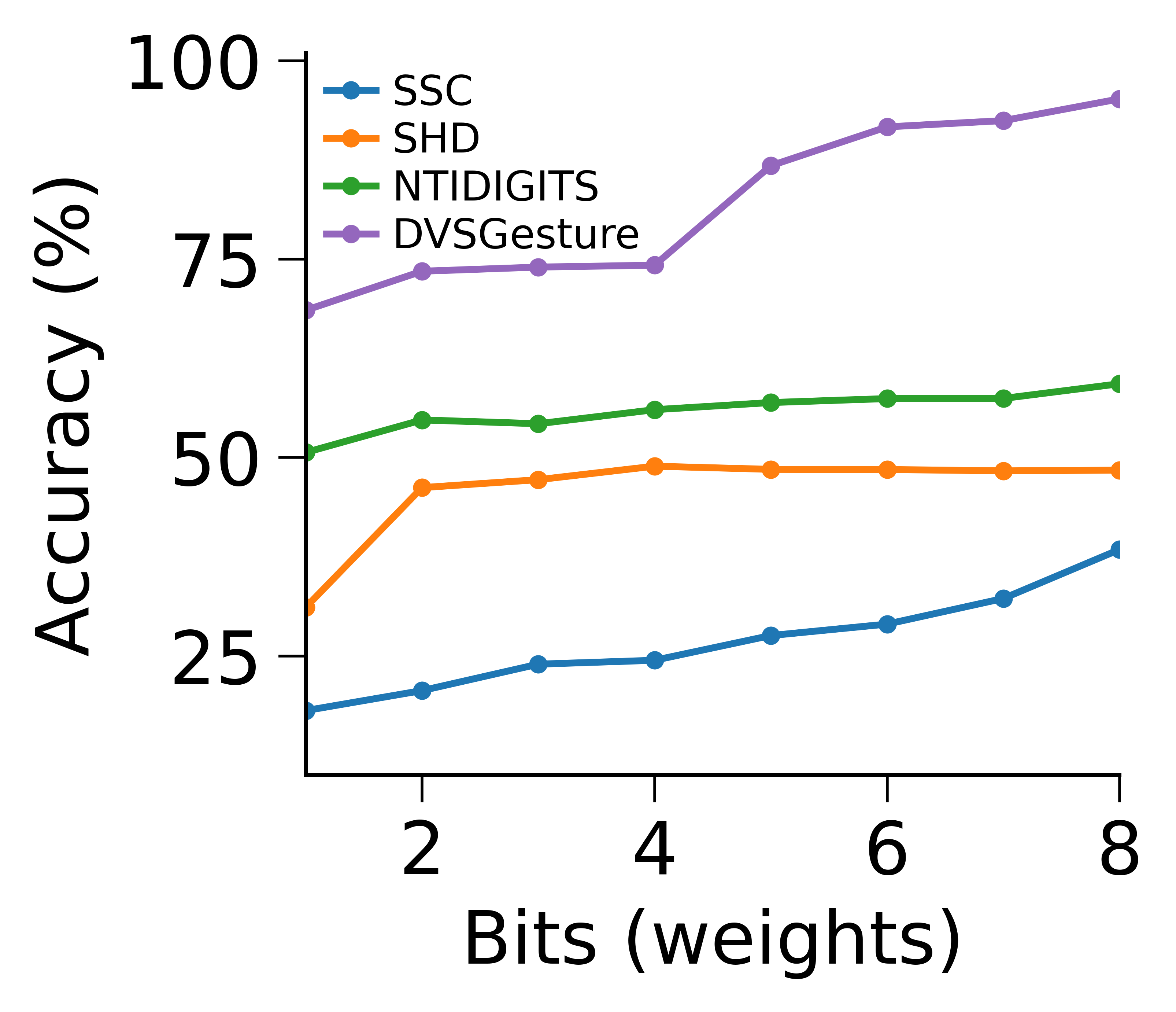}
\label{sup3_figure}
\end{figure}

We hypothesized that increasing synaptic weight precision alone would have limited impact on temporally rich tasks. Two feedforward spiking layers without axonal delays were trained while gradually increasing weight precision from 1 to 8 bits. Accuracy on SHD, SSC and NTIDIGITS rose modestly with higher precision, with gains plateauing quickly—SHD showed little improvement beyond 2 bits. This suggests that temporally complex tasks require additional mechanisms beyond weight precision to capture their dynamics. By contrast, DVS Gesture accuracy increased substantially with higher precision, consistent with its lower temporal complexity.
\clearpage
\suppfigsection{ Effect of neuron scaling with delays}{sup4} 
\begin{figure}[H]
\centering
\includegraphics[width=0.32\linewidth]{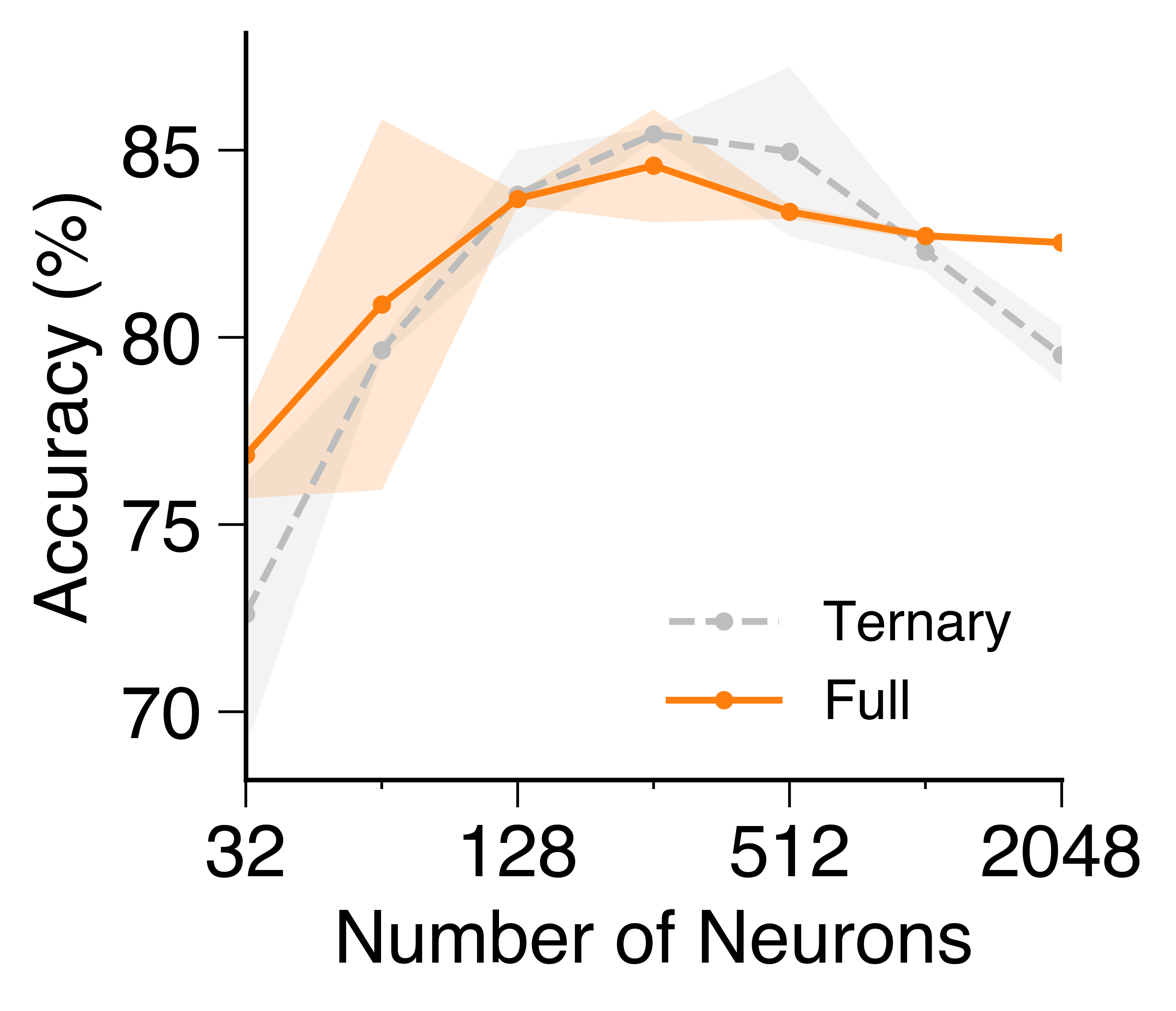}
\includegraphics[width=0.32\linewidth]{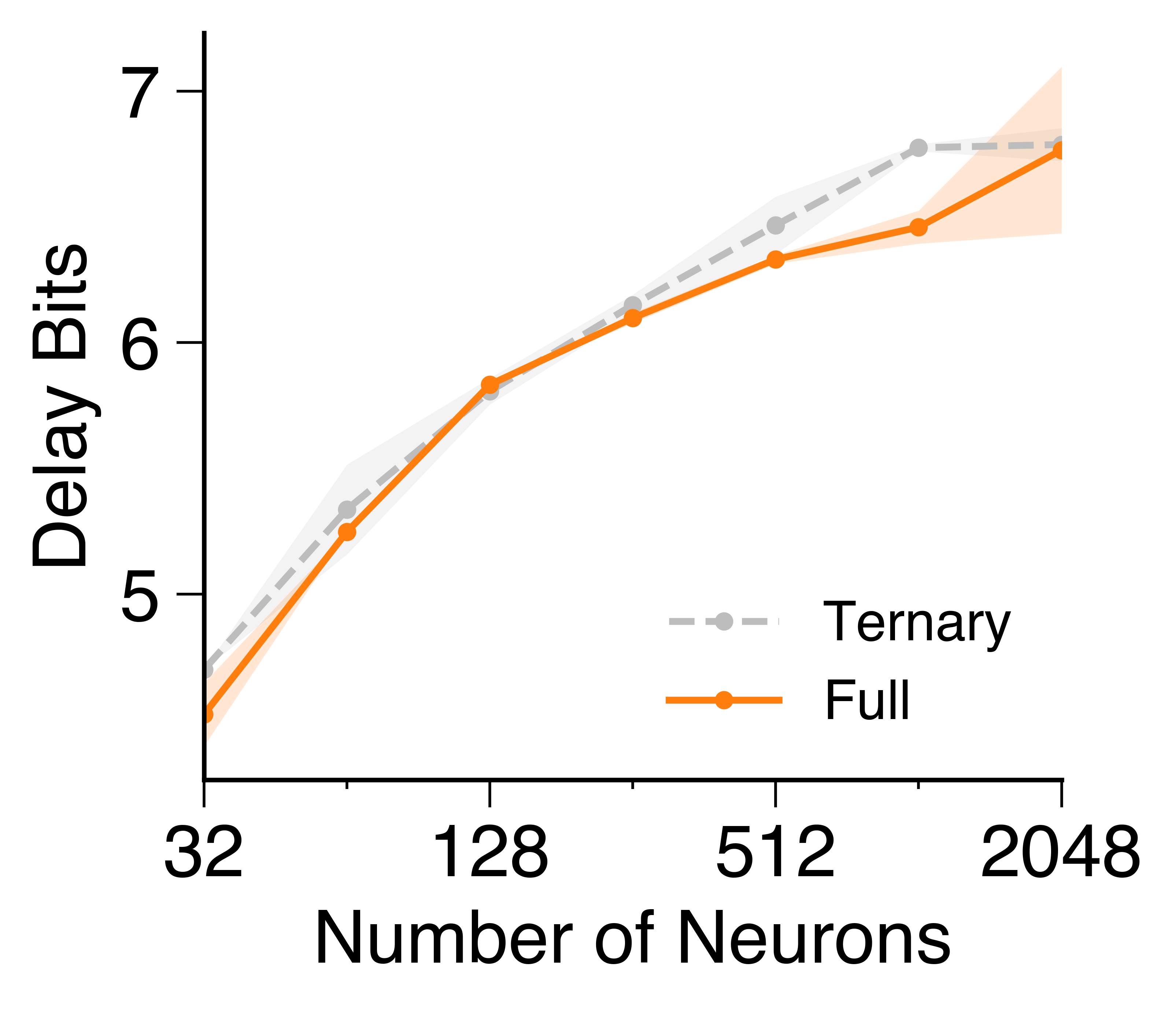}
\includegraphics[width=0.32\linewidth]{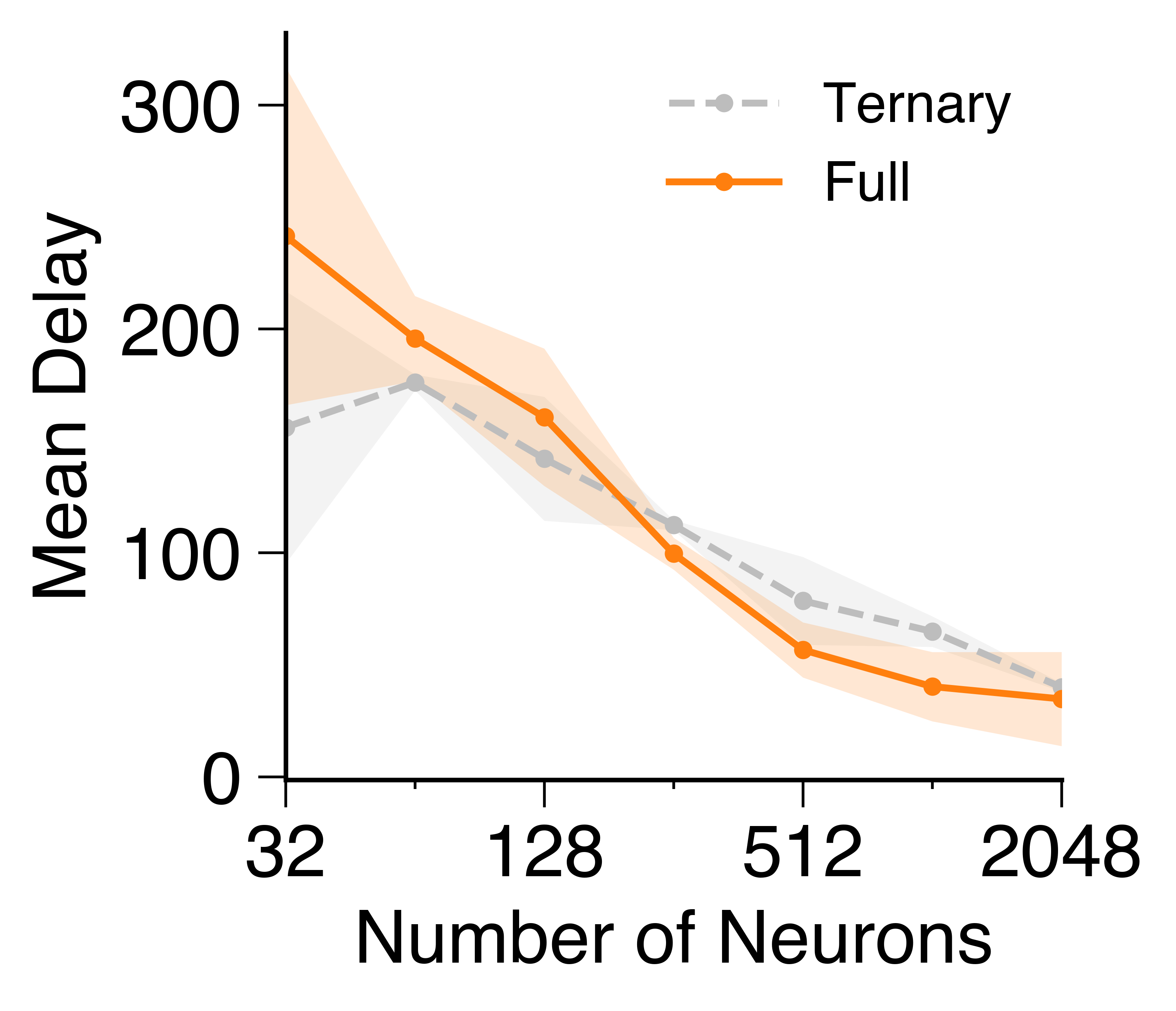}
\label{sup4_figure}
\end{figure}

We evaluated how increasing synaptic count interacts with axonal delays in a single-hidden-layer spiking network on SHD, comparing full-precision and ternary-precision weights. Accuracy increases rapidly as synapses are added, with diminishing returns at larger widths. This indicates that widening a feedforward layer alone does not reliably capture the task’s temporal structure under a fixed training setting. In contrast, the learned delay distributions systematically reorganise with scale: as synapse number increases, the mean axonal delay decreases while the set of distinct realised delay values expands, providing a richer temporal basis. This trend is monotonic across the tested range (the synapse axis is logarithmic) and is quantified by directly counting unique delay values. Together, these results suggest that performance gains track delay diversity more closely than network size alone.

\clearpage
\suppfigsection{ Relative contribution of delays to accuracy}{sup5} 
\begin{figure}[H]
\centering
\begin{subfigure}[]{0.32\linewidth}
  \caption{{SHD}}
\includegraphics[width=\linewidth]{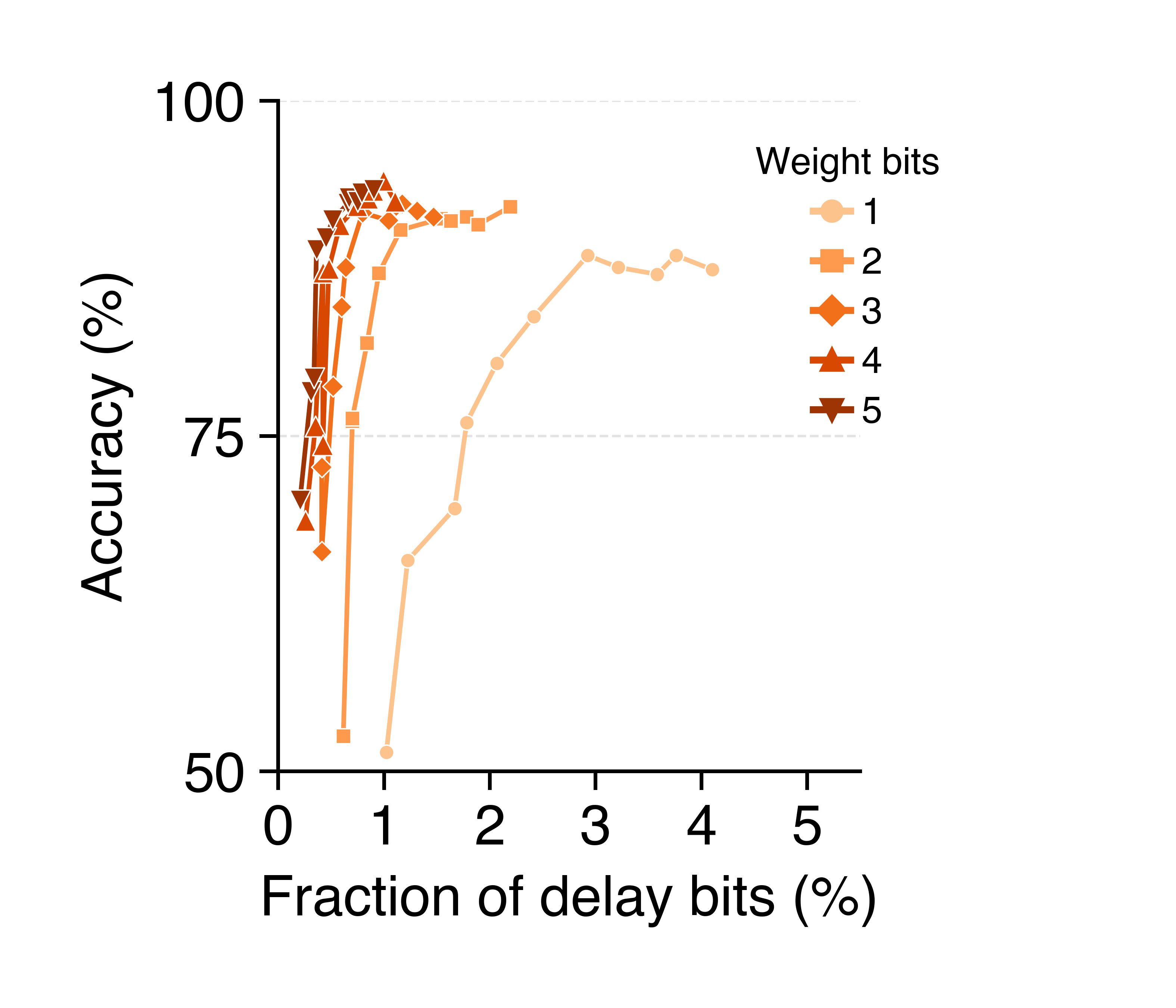}
\end{subfigure}
\begin{subfigure}[]{0.32\linewidth}
  \caption{{SSC}}
\includegraphics[width=\linewidth]{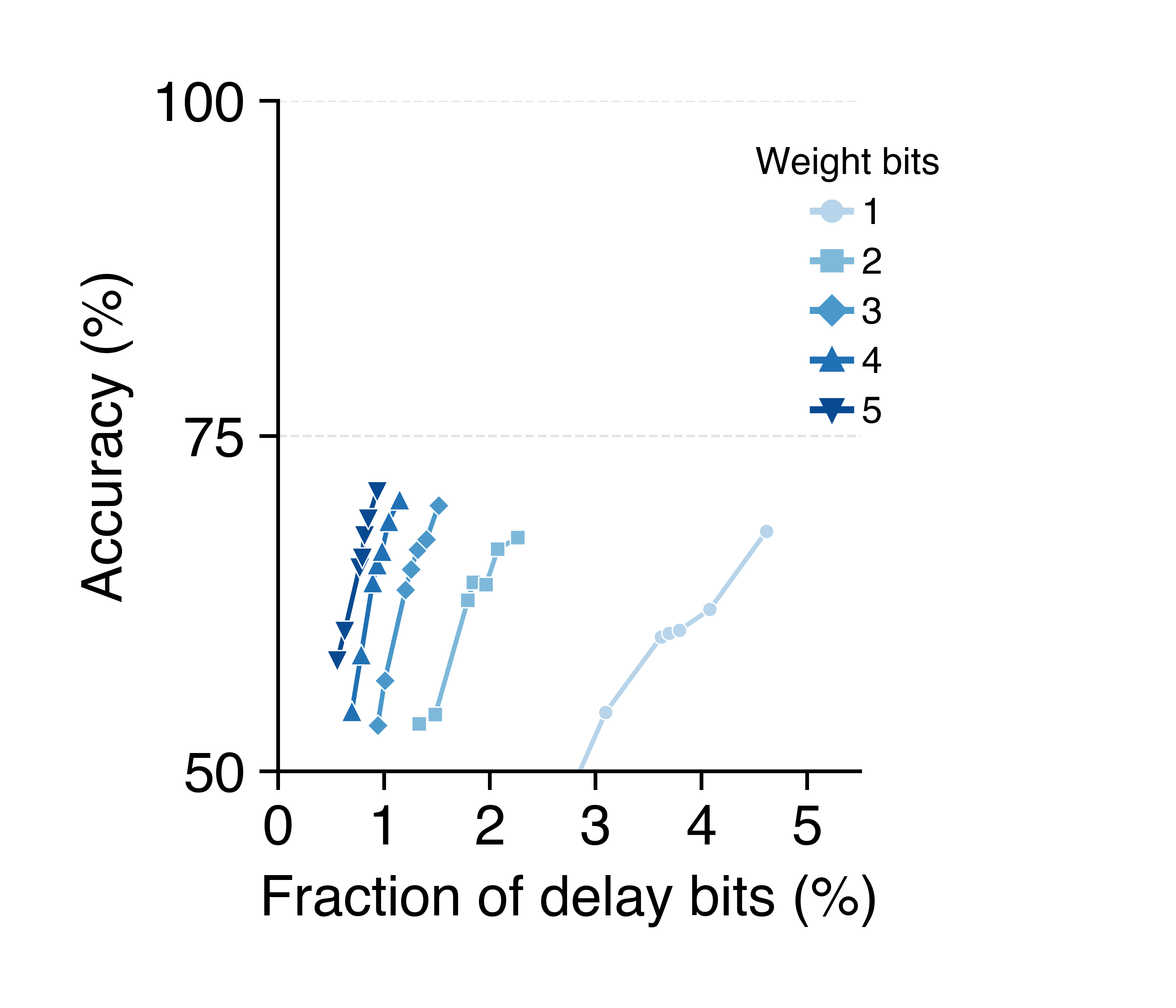}
\end{subfigure}
\begin{subfigure}[]{0.32\linewidth}
  \caption{{NTIDIGITS}}
\includegraphics[width=\linewidth]{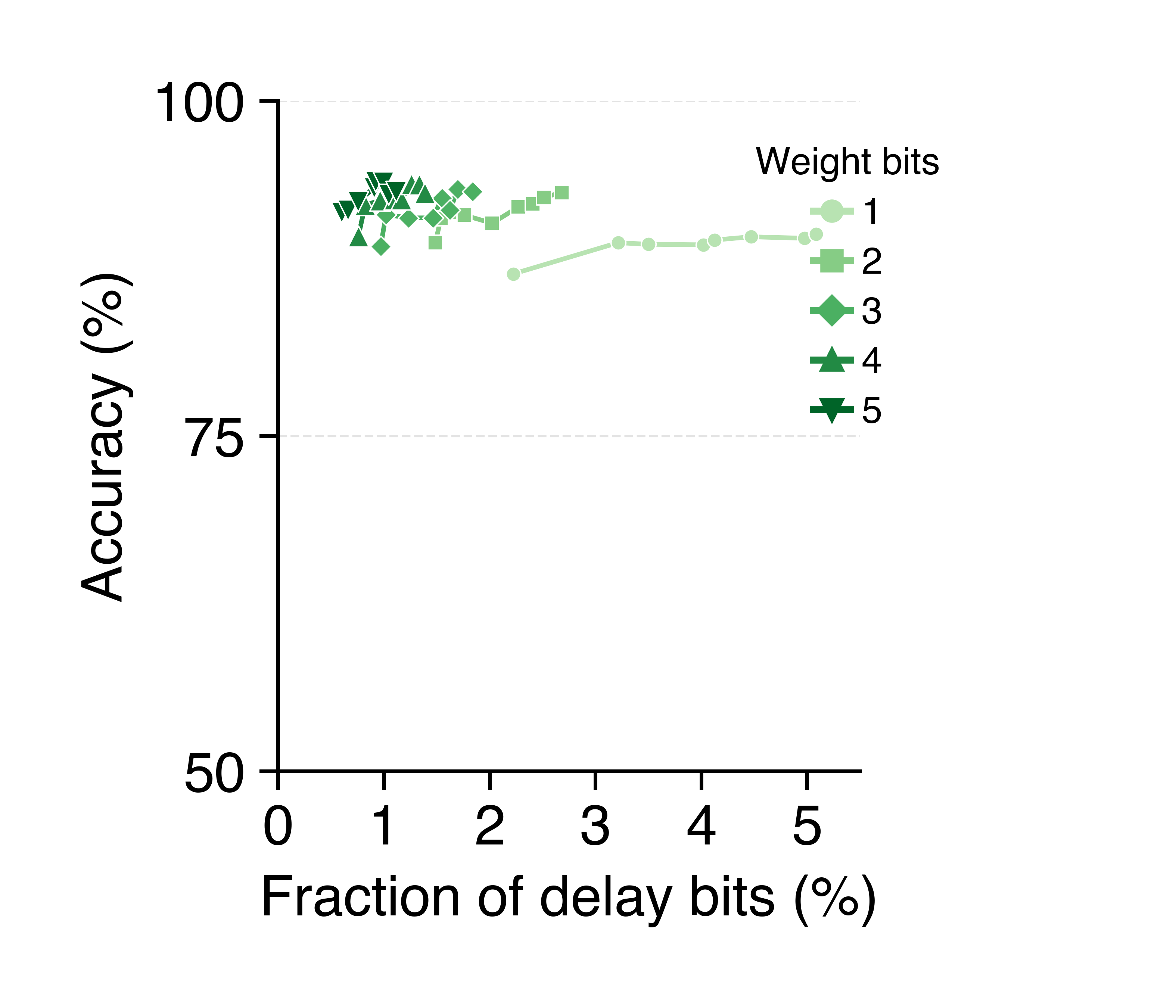}
\end{subfigure}
\label{sup5_figure}
\end{figure}

To quantify the relative contribution of delays, we define the delay-bit fraction
\begin{equation}
f_d = \frac{\mathrm{Bits}(\mathrm{delay})*n}{\mathrm{Bits}(\mathrm{delay})*n + \mathrm{Bits}(\mathrm{weight})*n^2}.
\end{equation}
Here, $n$ denotes the number of neurons (set to 128). For each fixed weight precision, we varied delay precision and measured accuracy. On SHD, accuracy rises steeply with $f_d$. On SSC, accuracy continues to improve with higher weight precision at comparable $f_d$, indicating that SSC is more constrained by synaptic precision than by delay resolution. On NTIDIGITS, performance remains sensitive to delay precision across weight settings.

\clearpage
\suppfigsection{Fixed random delay priors synergise with extreme weight quantisation}{weightdelaycodesign}
\begin{figure}[H]
\centering
\includegraphics[width=0.75\linewidth]{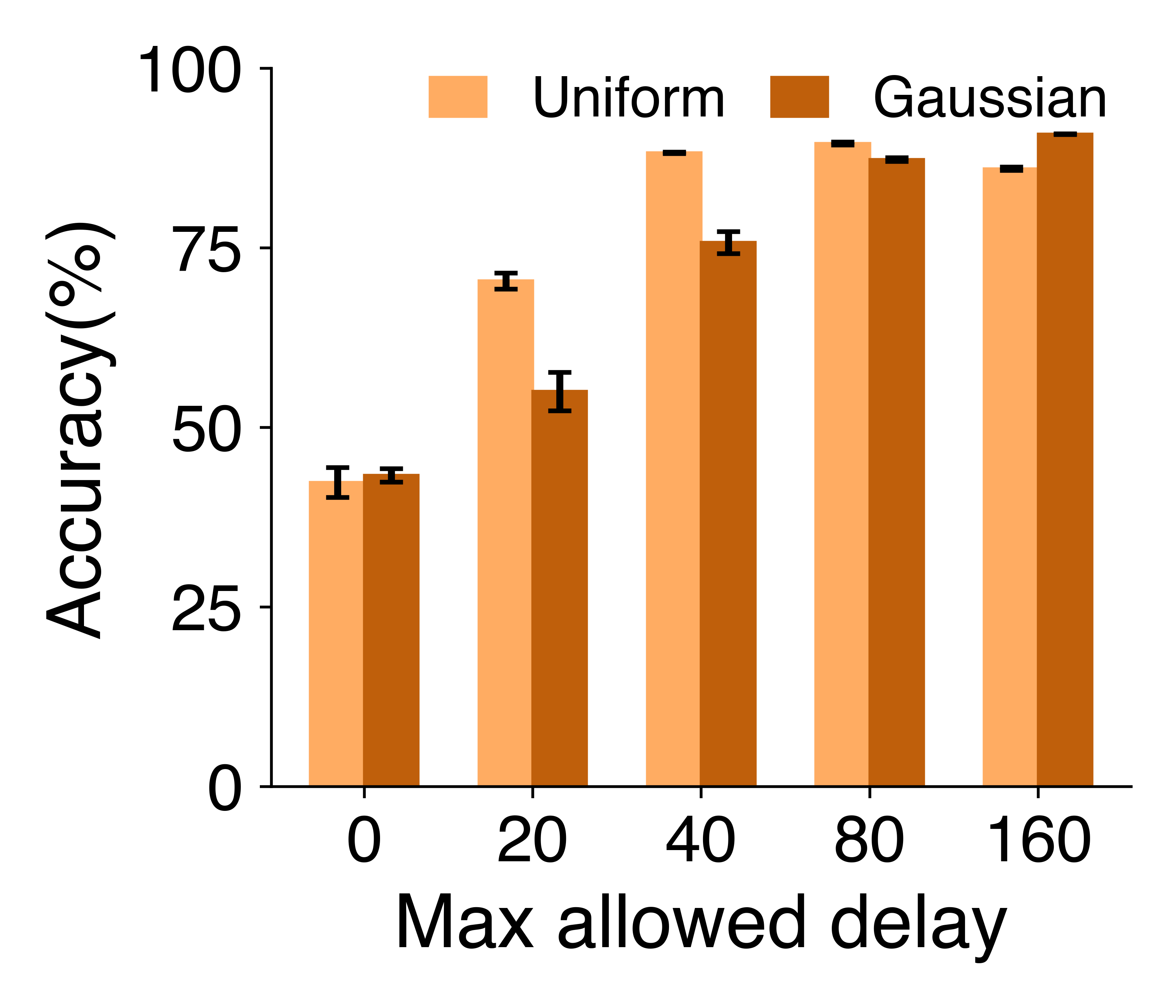}
\label{weightdelaycodesign_fig}
\end{figure}

{To decouple the contribution of temporal heterogeneity from delay optimisation, we fixed axonal delays by sampling them once from simple priors (uniform and Gaussian) and trained the network only with extreme learnable weight quantisation (1.58\,bits). For the uniform prior, each neuron’s delay was drawn independently as $d \sim \mathcal{U}(0,\theta_d)$, where $\theta_d$ denotes the max allowed delay. For the Gaussian prior, delays were sampled as $d \sim \mathcal{N}(\mu,\sigma^2)$ with $\mu=\theta_d/2$ and $\sigma=\theta_d/6$, then truncated to the valid range $[0,\theta_d]$ and discretised by rounding to integer time steps. In both cases, $\theta_d$ therefore directly controls the delay budget through the sampling support (uniform) or the truncation bound (Gaussian). Varying $\theta_d$ shows that even non-optimised heterogeneous delays can improve accuracy, with gains emerging once sufficiently long delays are permitted, consistent with a complementary division of labour between temporal offsets and low-bit synaptic structure. Nevertheless, these hand-specified priors do not match the performance obtained with trainable delays, indicating that the strongest improvements arise when delay distributions are shaped by data rather than prescribed a priori.}

\clearpage
\suppfigsection{Effects of long and short delays with delay pruning} {sup6}
\begin{figure}[H]
\centering
\begin{subfigure}[]{0.45\linewidth}
  \caption{{SHD}}
  \includegraphics[width=\linewidth]{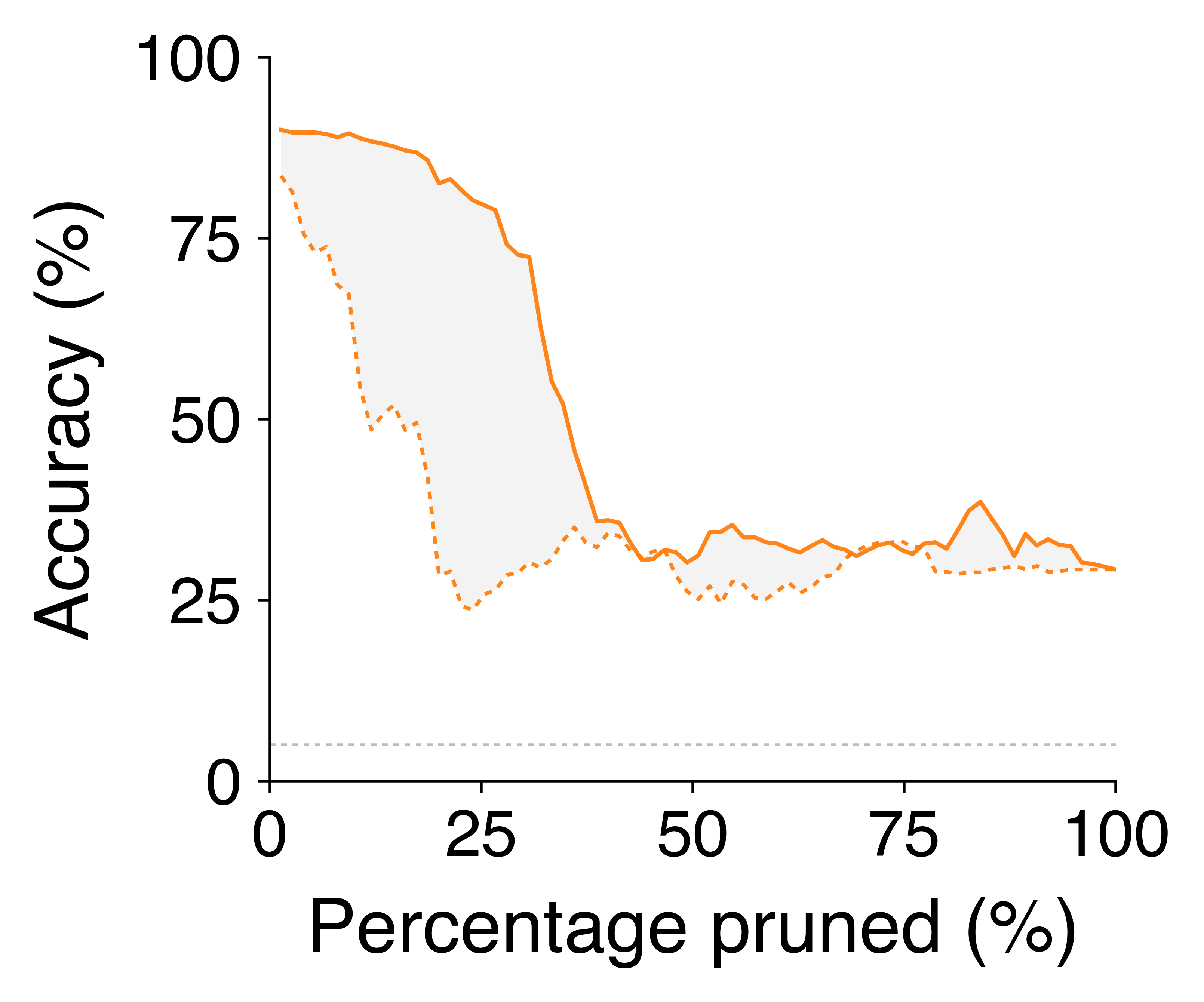}
\end{subfigure}
\begin{subfigure}[]{0.45\linewidth}
  \caption{SSC}
  \includegraphics[width=\linewidth]{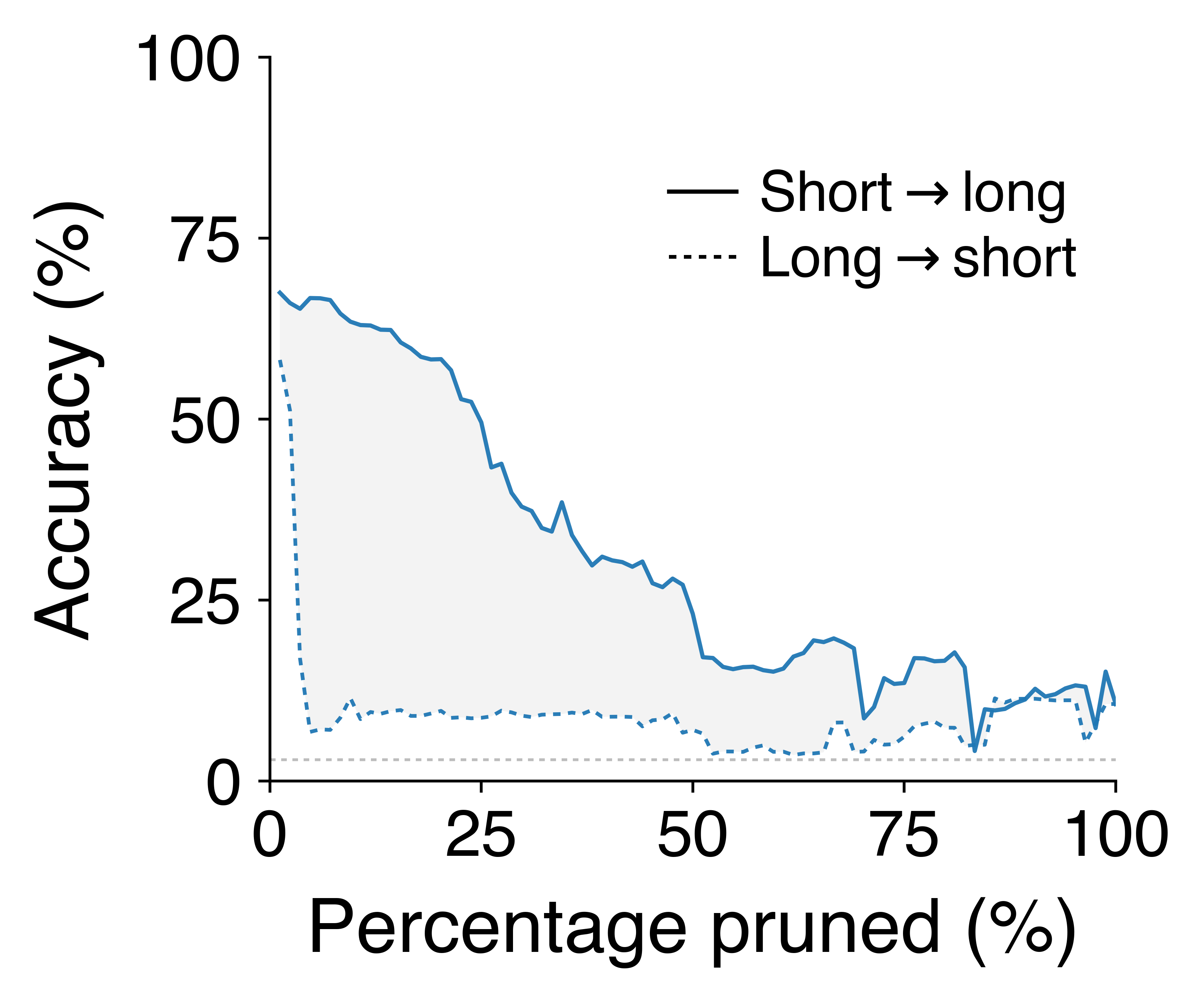}
\end{subfigure}

\begin{subfigure}[]{0.45\linewidth}
  \caption{{NTIDIGITS}}
  \includegraphics[width=\linewidth]{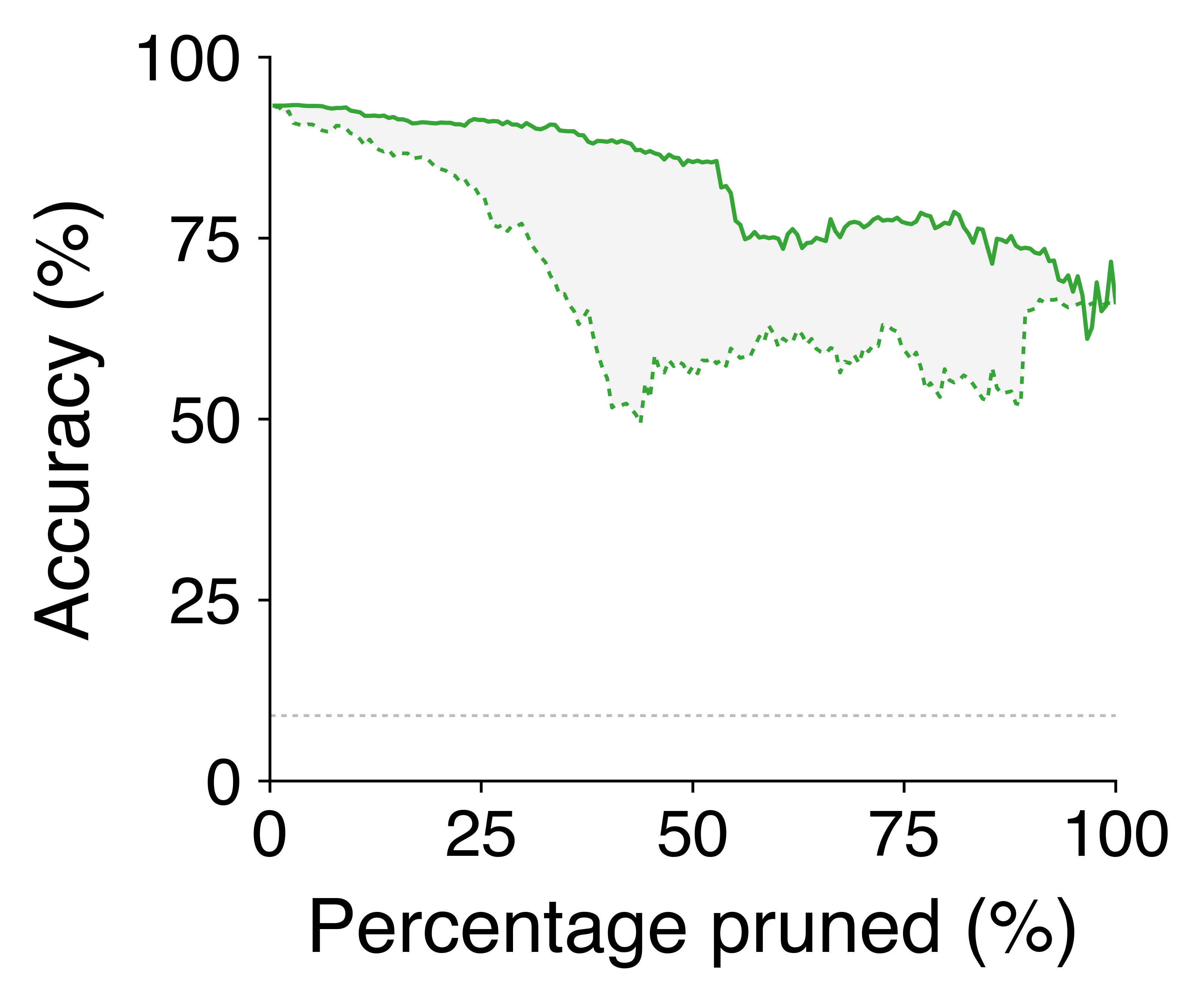}
\end{subfigure}
\begin{subfigure}[]{0.45\linewidth}
  \caption{{DVS-Gesture}}
  \includegraphics[width=\linewidth]{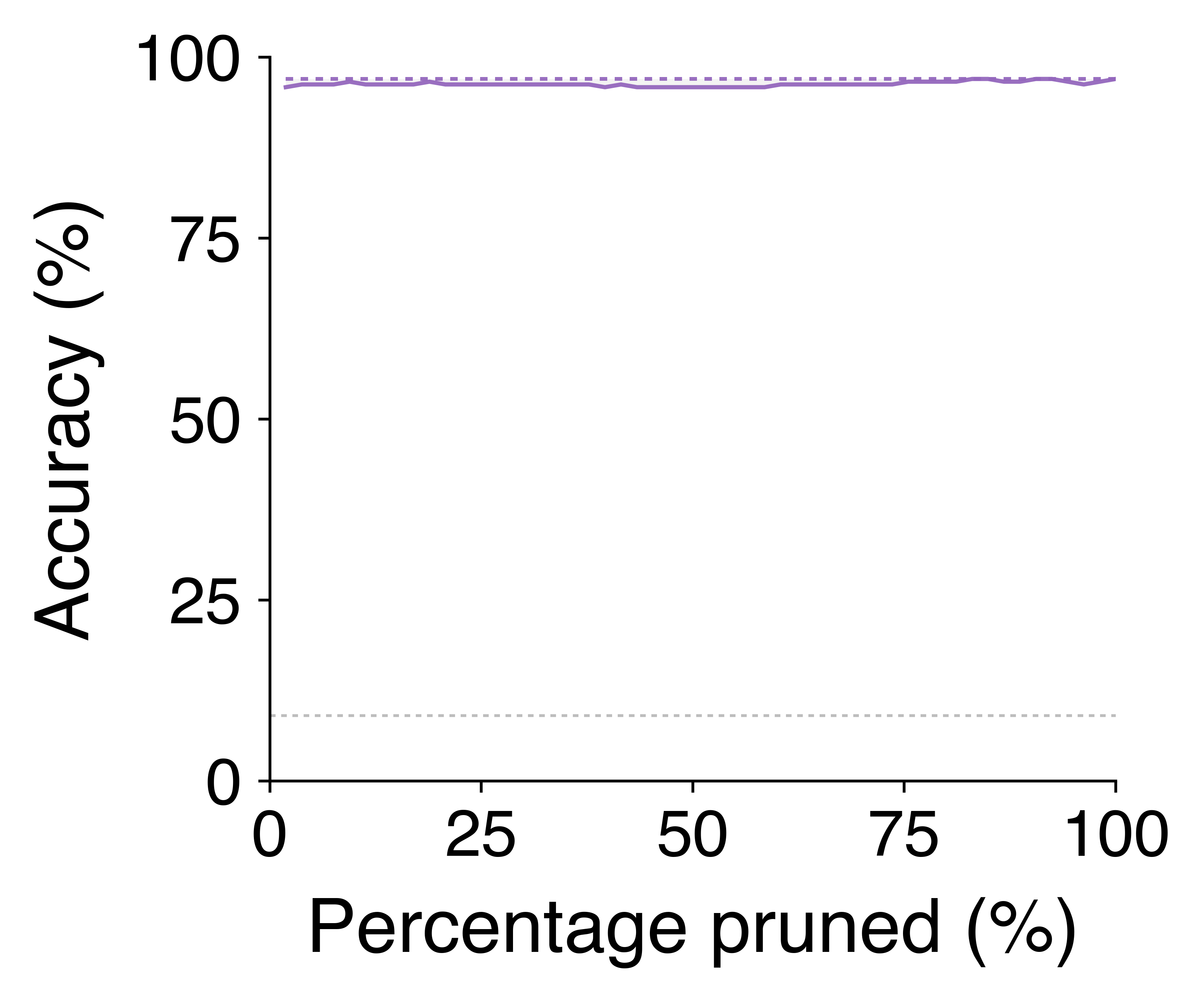}
\end{subfigure}

\label{sup6_figure}
\end{figure}
We performed systematic ablations by pruning delays in order of magnitude—short-to-long and long-to-short. Zeroing short delays had little effect, whereas eliminating even a few long delays precipitated a steep accuracy collapse. Pruning only the two longest delays reduced accuracy below 80\%, 17\%, and 92\% for SHD, SSC and NTIDIGITS, respectively; zeroing the top 10\% of delays further depressed performance to approximately 50\%, 10\%, and 87\% for the same datasets. An instructive exception was DVS Gesture, for which delay removal had negligible impact. Notably, even after complete delay ablation the networks retained partial classification ability across all tasks, indicating that weights alone can exploit spike-rate information \cite{yu2025beyond}.

\clearpage
\suppfigsection{Effect of constraining the maximum axonal delay during training}{supdelaymax}
\begin{figure}[H]
\centering
\includegraphics[width=0.75\linewidth]{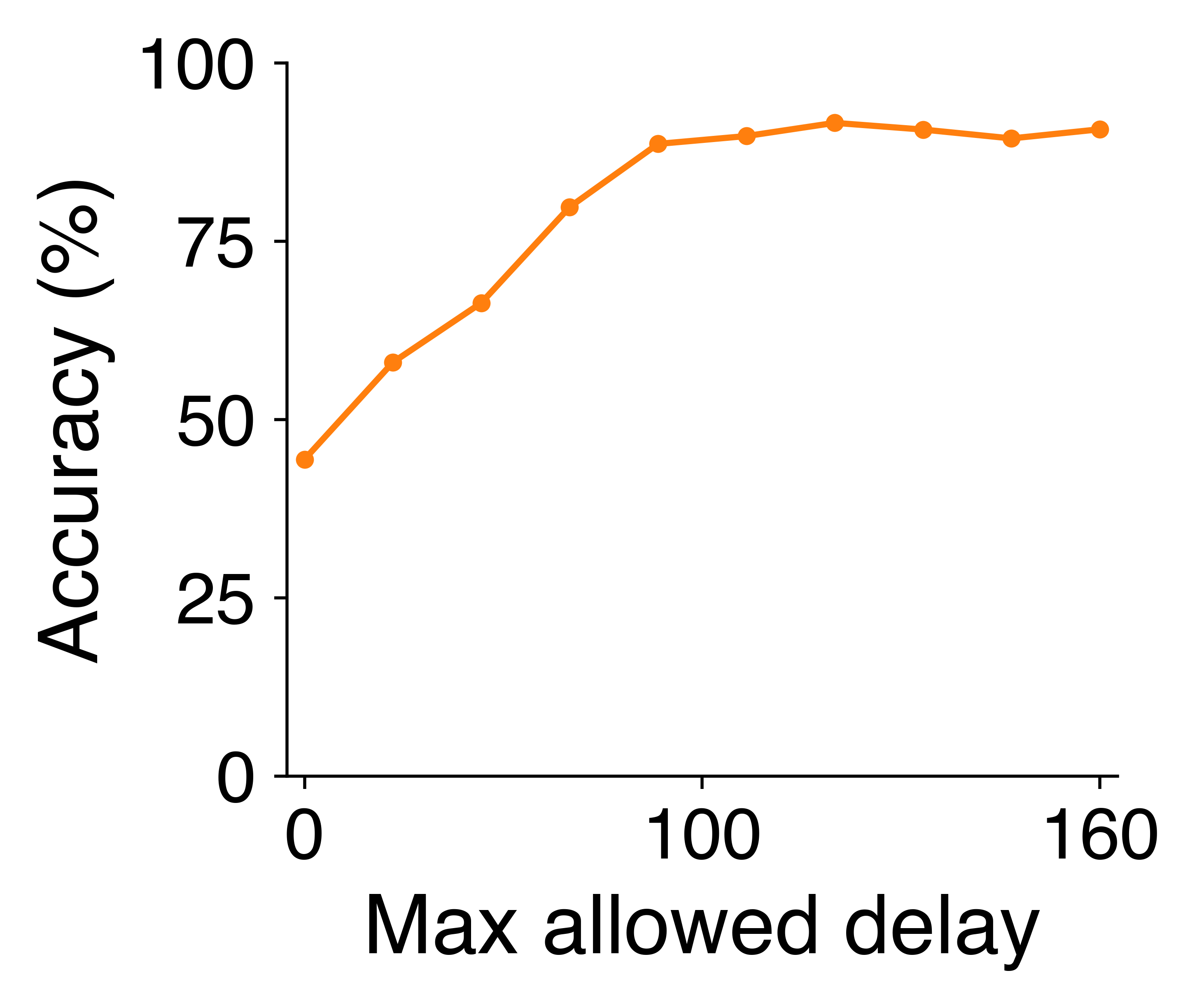}
\label{supdelaymax1}
\end{figure}

{We trained the network under an explicit maximum-delay constraint by clamping learnable axonal delays to $d \in [0,\theta_d]$ throughout training, where $\theta_d$ is the max allowed delay. On SHD, relaxing this constraint yields a steep accuracy gain that saturates once sufficiently long delays are permitted, while further increases provide little additional benefit and can slightly reduce performance. This pattern might indicate that access to long delays is necessary to capture long-range temporal dependencies in this temporally structured task.}

\clearpage
\suppfigsection{ Effect of delay distribution with longer time constant} {sup7}
\begin{figure}[H]
\centering

\begin{subfigure}[]{0.45\linewidth}
  \caption{$\tau=1$}
  \includegraphics[width=\linewidth]{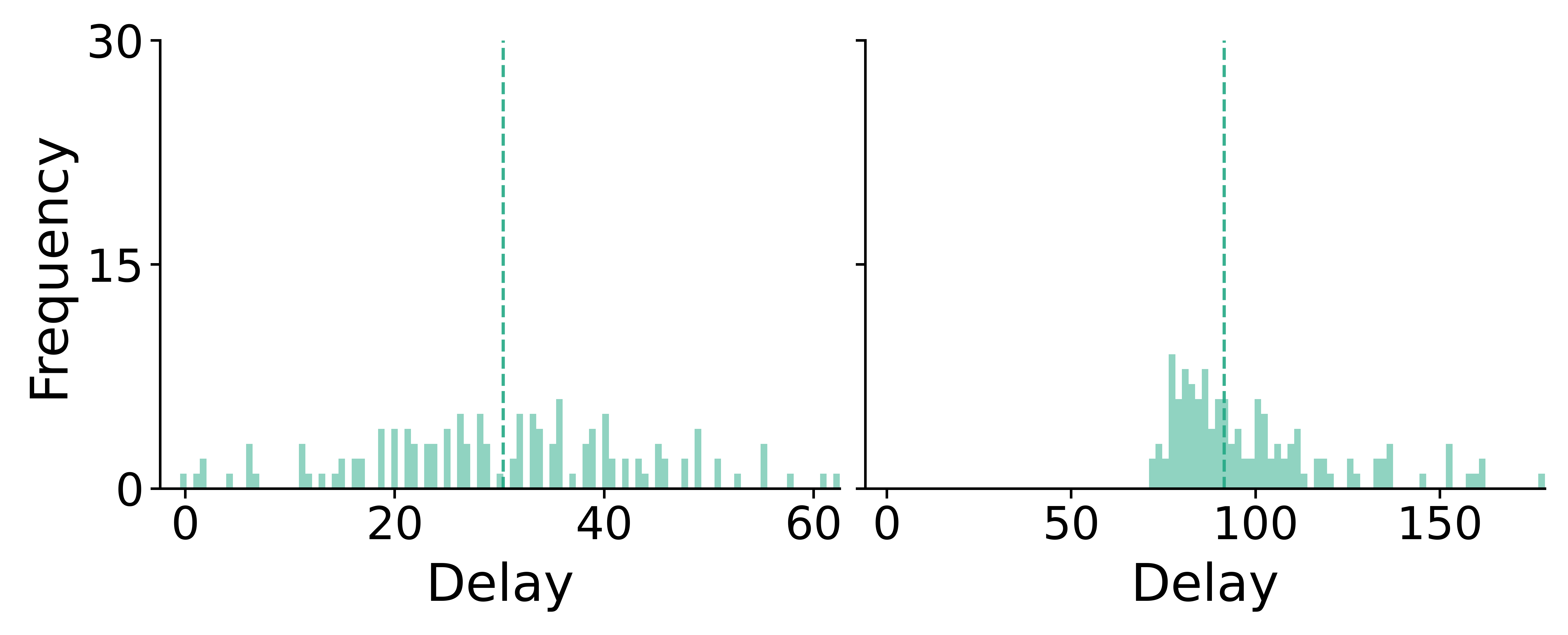}
\end{subfigure}
\begin{subfigure}[]{0.45\linewidth}
  \caption{$\tau=2$}
  \includegraphics[width=\linewidth]{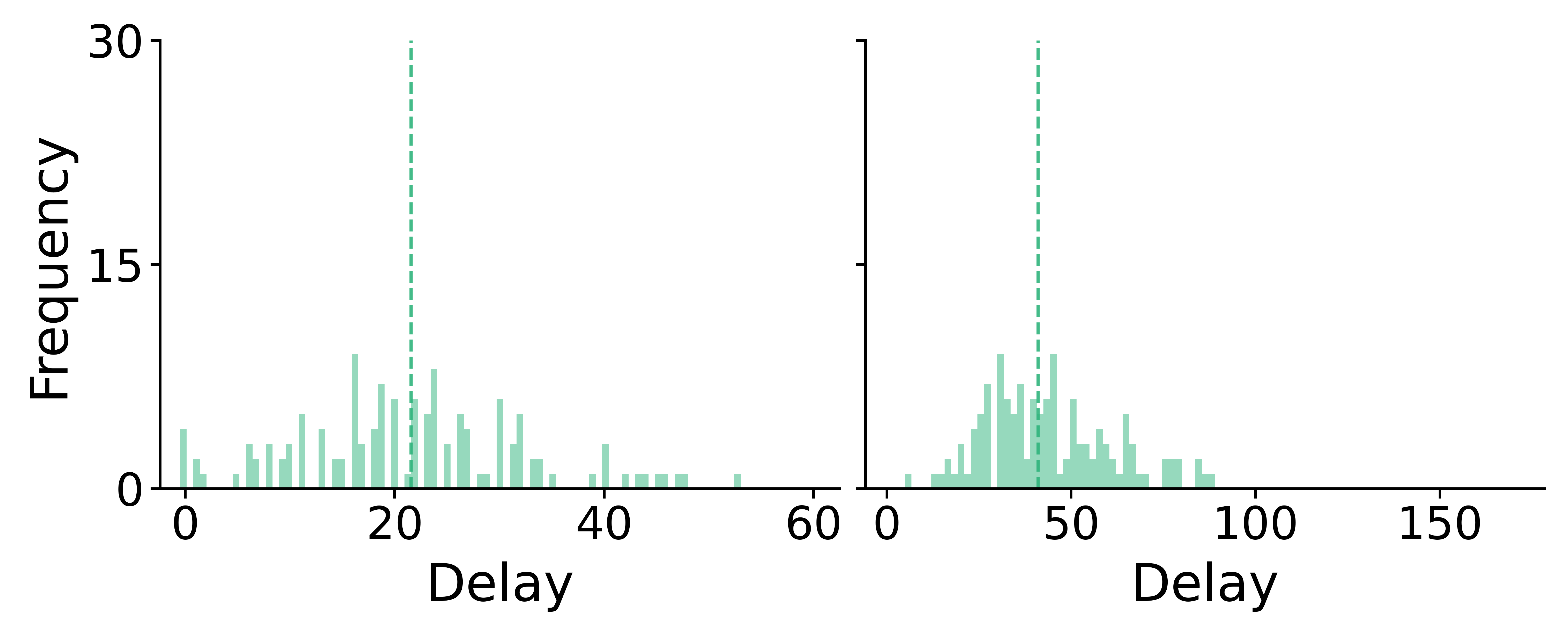}
\end{subfigure}

\begin{subfigure}[]{0.45\linewidth}
  \caption{$\tau=3$}
  \includegraphics[width=\linewidth]{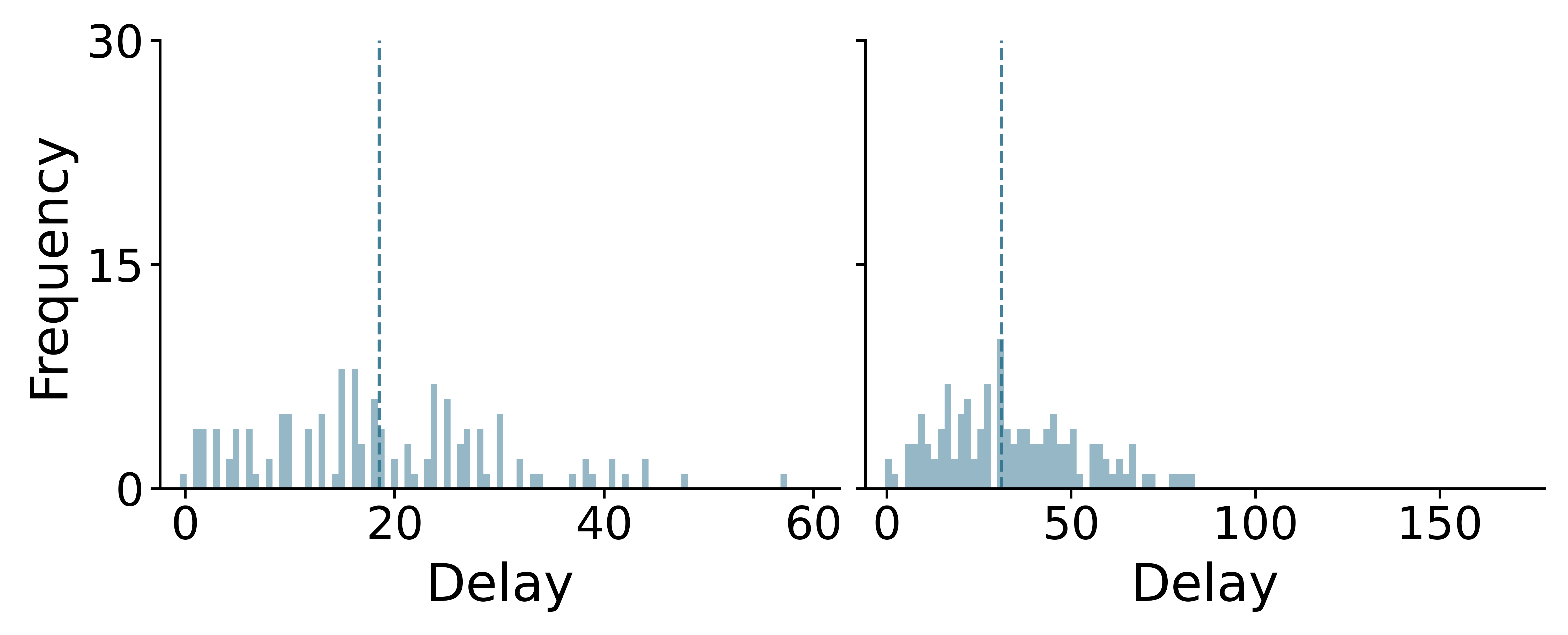}
\end{subfigure}
\begin{subfigure}[]{0.45\linewidth}
  \caption{$\tau=4$}
  \includegraphics[width=\linewidth]{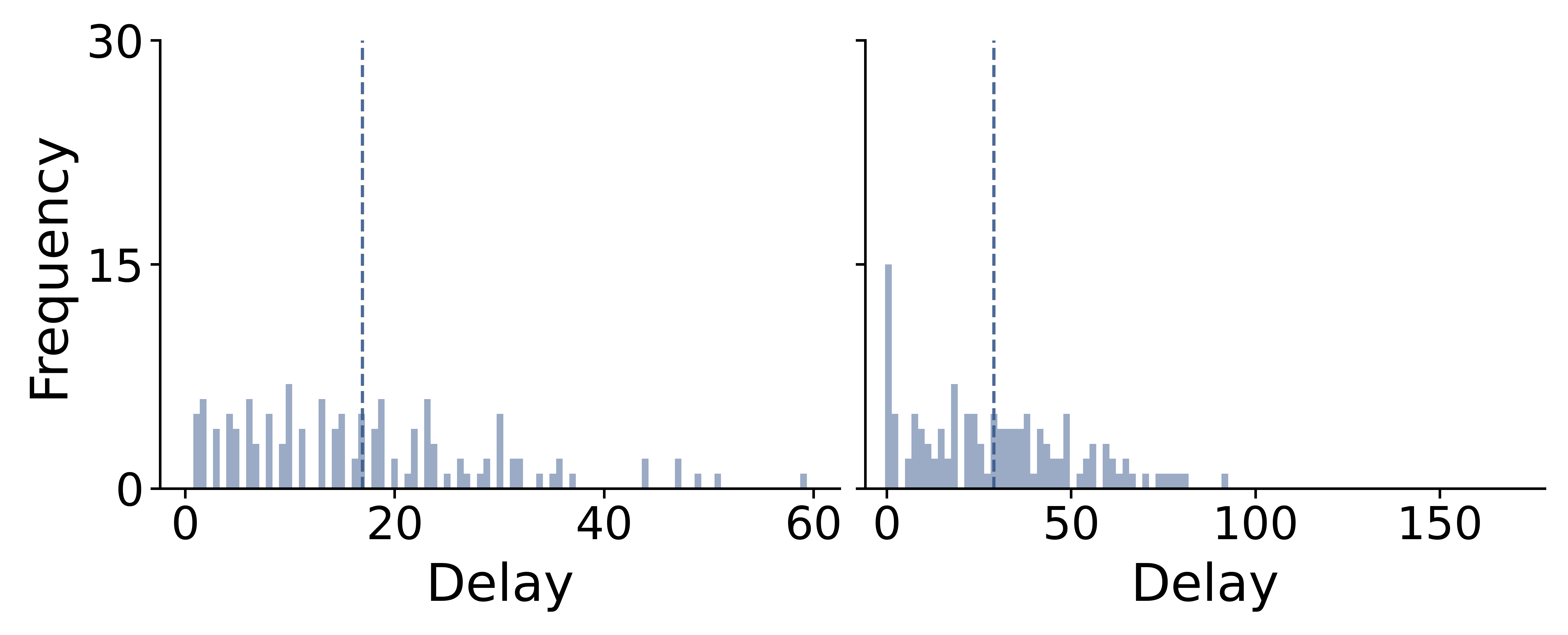}
\end{subfigure}

\begin{subfigure}[]{0.45\linewidth}
  \caption{$\tau=5$}
  \includegraphics[width=\linewidth]{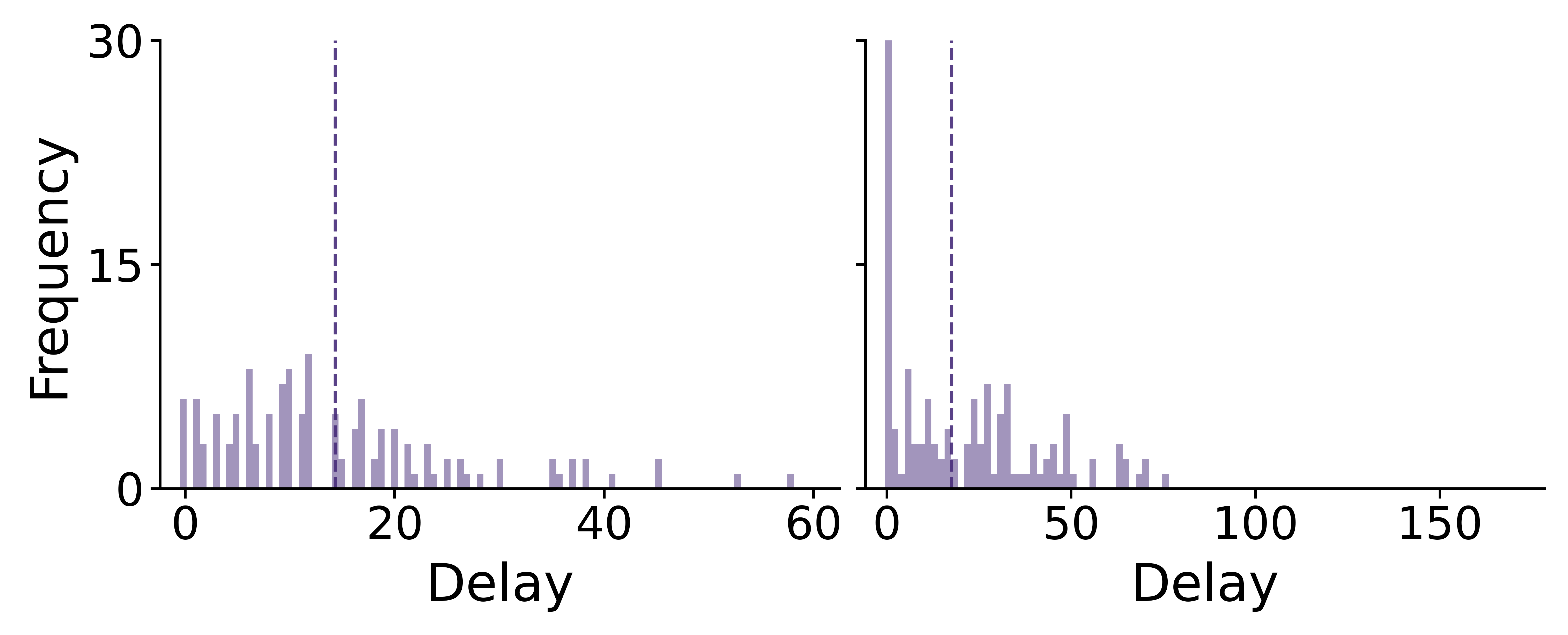}
\end{subfigure}

\label{sup7_figure}
\end{figure}
Longer time constants ($\tau$) partially absorb the temporal span that would otherwise be handled by axonal delays. The panels depict the learned delay distributions as $\tau$ increases ($\tau\!=\!1$ to $5$) with all other settings fixed and delays trained. Increasing $\tau$ shifts the mass toward shorter delays and reduces dispersion, yet a non-trivial long tail persists, most prominently in later layers, indicating that sparse long delays remain critical for aligning temporally separated features. For each $\tau$, the left and right panels correspond to the first and second hidden layers, respectively; dashed lines indicate the layer-wise mean delay.

\clearpage

 \section*{Supplementary Tables}
\captionsetup[table]{labelformat=default}

\captionsetup{type=table}
\renewcommand{\thetable}{S\arabic{table}}
\setcounter{table}{0}
\captionsetup[table]{name=Table}

\begin{center}
\caption{\textbf{Test accuracy across datasets under full precision and delay conditions.}
Results are reported as percentages. Each condition combines full synaptic weight precision with either the presence or absence of trainable axonal delays. Chance-level accuracies are shown in the final row for reference.}
\label{tab:accuracy}
\begin{tabular}{llcccc}
  \toprule
  \textbf{Precision} & \textbf{Delays} & \textbf{SHD} & \textbf{SSC} & \textbf{NTIDIGITS} & \textbf{DVS Gesture} \\
  \midrule
  Full precision & No  & 48.60 & 38.50 & 60.02 & 93.56 \\
  Full precision & Yes & 90.98 & 70.94 & 93.28 & 95.83 \\
  \midrule
  \multicolumn{2}{l}{Chance level} & 5.00 & 2.90 & 9.00 & 9.00 \\
  \bottomrule
\end{tabular}
\end{center}

\clearpage
\captionsetup{type=table}
\renewcommand{\thetable}{S\arabic{table}}
\setcounter{table}{1}
\captionsetup[table]{name=Table}
\begin{center}
\captionof{table}{{\textbf{Test accuracy on SHD under excitatory/inhibitory sign constraints and delay learning.}
All conditions restrict each layer to a single positive weight value and zero. We vary whether negative weights are permitted (none; readout only; all layers), and report the corresponding accuracy with learned axonal delays. Chance-level accuracy is shown in the final row for reference.}}
\label{excitatory}

\begin{adjustbox}{max width=\linewidth}
\begin{tabular}{l l l r}
\toprule
\textbf{Excitatory-only} & \textbf{Inhibitory allowed} & \textbf{Delays} & \textbf{Accuracy (\%)} \\
\midrule
$\checkmark$ & $\times$ & No  & 40.54 \\
$\checkmark$ & $\times$ & Yes & 50.08 \\
$\checkmark$ & $\checkmark$ (readout layer only) & Yes & 87.72 \\
$\checkmark$ & $\checkmark$ & Yes & 90.98 \\
\midrule
\multicolumn{3}{l}{Chance level} & 5.00 \\
\bottomrule
\end{tabular}
\end{adjustbox}
\end{center}

\end{document}